\documentclass[preprint]{article}

\usepackage{arxiv}

\usepackage[utf8]{inputenc} % allow utf-8 input
\usepackage[T1]{fontenc}    % use 8-bit T1 fonts
\usepackage{hyperref}       % hyperlinks
\usepackage{url}            % simple URL typesetting
\usepackage{booktabs}       % professional-quality tables
\usepackage{amsfonts}       % blackboard math symbols
\usepackage{nicefrac}       % compact symbols for 1/2, etc.
\usepackage{microtype}      % microtypography
\usepackage{lipsum}
\usepackage{graphicx}

\usepackage{framed,multirow}
\usepackage{amsmath,amssymb,amsthm}
\usepackage{algorithmic}
\usepackage[algo2e,ruled,vlined]{algorithm2e}
\usepackage{upgreek}
\usepackage{latexsym}
\usepackage{url}
\usepackage{xcolor}
\usepackage{hyperref}
\usepackage{booktabs}
\usepackage{parskip}
\definecolor{newcolor}{rgb}{.8,.349,.1}

\newtheorem{theorem1}{Theorem}
\usepackage{multirow}
\usepackage{pifont}
\newcommand{\cmark}{\ding{51}}%
\newcommand{\xmark}{\ding{55}}%

\renewcommand{\vec}[1]{\mathbf{#1}}
\newcommand{\mr}[1]{\mathrm{#1}}
\newcommand{\tx}[1]{\textrm{#1}}
\newcommand{\signif}[1]{\underline{#1}}

\newcommand{\phm}{\phantom{0}}
\newcommand{\B}{\bfseries}
\newcommand{\viz}[1]{\includegraphics[width=0.107\textwidth,height=0.107\textwidth]{Figures/#1.png}}
\newcommand{\alphaJSD}[1]{\includegraphics[width=0.25\textwidth]{Figures/alphaJSD/alpha#1.png}}
\newcommand{\entropyMaps}[1]{\includegraphics[width=0.16\textwidth,height=0.165\textwidth]{Figures/entropyMaps/#1.png}}

\newcommand{\mypar}[1]{\noindent\paragraph*{\textbf{#1}}}
\newcommand{\JSD}{\mr{JSD}}
\newcommand{\kl}[2]{D_{\mr{KL}}\big(#1 \, || \, #2\big)}
\newcommand{\loss}{\ell}
\newcommand{\Loss}{\mathcal{L}}
\newcommand{\lab}{\mathcal{S}}
\newcommand{\unlab}{\mathcal{U}}

\newcommand{\pp}{\vec{p}}
\newcommand{\img}{\vec{x}}
\newcommand{\gt}{\vec{y}}
\newcommand{\classes}{\mathcal{C}}
\newcommand{\pslab}{\widehat{y}}

\newcommand{\params}{\boldsymbol{\uptheta}}
\newcommand{\pixels}{\Omega}
\newcommand{\ZZ}{\mathbb{Z}}
\newcommand{\Real}{\mathbb{R}}
\newcommand{\entr}{\mathcal{H}}

\newcommand{\reg}{\mathcal{R}}
 
\newcommand{\ww}{\vec{w}} 
\newcommand{\LossSPCT}{\mathcal{L}}

\newcommand{\ppi}{\boldsymbol{\pi}}

\newcommand{\modifR}[1]{#1}

\title{Self-paced and self-consistent co-training for semi-supervised image segmentation}

\author{
  Ping Wang\thanks{Corresponding author} \\
  Department of Software and IT Engineering\\ Ecole de technologie sup\'erieure\\
  Montreal, H3C1K3, Canada \\
  \texttt{ping.wang.1@ens.etsmtl.ca} \\
   \And
 Jizong Peng \\
  Department of Software and IT Engineering\\ Ecole de technologie sup\'erieure\\
  Montreal, H3C1K3, Canada \\
  \texttt{jizong.peng.1@etsmtl.net} \\
  \And
  Marco Pedersoli \\
  Department of Systems Engineering\\ 
  Ecole de technologie sup\'erieure\\
  Montreal, H3C1K3, Canada \\
  \texttt{marco.pedersoli@etsmtl.ca} \\
  \And
  Yuanfeng Zhou \\
  School of Computer Science and Technology\\
  Shandong University\\ 
  Jinan, 250101, China\\
  \texttt{yfzhou@sdu.edu.cn} \\
  \And
  Caiming Zhang \\
  School of Computer Science and Technology\\
  Shandong University\\ 
  Jinan, 250101, China\\
  \texttt{czhang@sdu.edu.cn} \\
  \And
  Christian Desrosiers \\
  Department of Software and IT Engineering\\ 
  Ecole de technologie sup\'erieure\\
  Montreal, H3C1K3, Canada \\
  \texttt{christian.desrosiers@etsmtl.ca} \\
}

\begin{document}
\maketitle

\begin{abstract}
Deep co-training has recently been proposed as an effective approach for image segmentation when annotated data is scarce. In this paper, we improve existing approaches for semi-supervised segmentation with a self-paced and self-consistent co-training method. To help distillate information from unlabeled images, we first design a self-paced learning strategy for co-training that lets jointly-trained neural networks focus on easier-to-segment regions first, and then gradually consider harder ones. This is achieved via an end-to-end differentiable loss in the form of a generalized Jensen Shannon Divergence (JSD). Moreover, to encourage predictions from different networks to be both consistent and confident, we enhance this generalized JSD loss with an uncertainty regularizer based on entropy. The robustness of individual models is further improved using a self-ensembling loss that enforces their prediction to be consistent across different training iterations. We demonstrate the potential of our method on three challenging image segmentation problems with different image modalities, using small fraction of labeled data. Results show clear advantages in terms of performance compared to the standard co-training baselines and recently proposed state-of-the-art approaches for semi-supervised segmentation.
\end{abstract}

% keywords can be removed
%\keywords{Co-training \and Image segmentation  \and Self-paced learning \and Semi-supervised learning \and Temporal ensembling}

\section{Introduction}
Semi-superived learning, where the goal is to learn a given task with few labeled examples and many unlabeled ones, has generated growing interest in research. This learning paradigm is of key importance for medical imaging~\citep{cheplygina2019not} since obtaining annotated data in applications of this field is often costly and typically requires highly-trained experts. In the last years, a broad range of deep learning approaches have been proposed for semi-supervised learning. A method which has recently gained popularity is co-training. Initially proposed by Blum and Mitchell for classification~\citep{blum1998combining}, this method exploits the idea that training examples can often be described by two complementary sets of features called views. It was shown that models trained collaboratively on conditionally-independent views improve semi-supervised performance with PAC-style bounds on the generalization error~\citep{dasgupta2002pac}.

The standard dual-view co-training approach consists in learning a separate classifier for each view using labeled data. Information is then exchanged between classifiers based on their high-confidence predictions for unlabeled data. The generalization of this approach to more than two views is called multi-view learning~\citep{xu2013survey}. By enforcing agreement between classifiers, multi-view learning constrains the parameter search space and helps find models which can generalize to unseen data. Despite its success in various classification tasks, its application to image analysis problems has so far been limited, mostly due to the requirement of having independent features in each view. Although these complementary views are available in specific scenarios such as multiplanar images~\citep{zhou2019semi,xia20203d}, there is no effective way to construct them for arbitrary images. \cite{qiao2018deep} proposed a deep co-training method for semi-supervised image recognition, where adversarial examples built from training images are used to enforce diversity among different classifiers. This idea was later extended to medical image segmentation by \cite{peng2019deep}. More recently, \cite{xia20203d} introduced an uncertainty-aware multi-view co-training framework in which prediction uncertainty estimated using a Bayesian approach is employed to merge the output for different views. 

Despite improving performance when few labeled images are available, existing co-training approaches for semi-supervised segmentation suffer from two important limitations: 1) they do not employ a self-paced learning strategy and therefore are susceptible to incorrect predictions in initial stages of training; 2) they do not exploit self-consistency within the model. In this paper, we address these limitations by introducing a self-paced and self-consistent co-training method. The main contributions of our work are as follows:
\begin{itemize}\setlength\itemsep{0em}
 \item We propose, to our knowledge, the first self-paced co-training method for semi-supervised segmentation. We show that the proposed self-paced learning strategy corresponds to minimizing a generalized Jensen-Shannon Divergence (JSD), where the confidence of individual models for each pixel in unlabeled images is considered. This enables the learning process to focus on transferring high-confidence predictions across models, thereby boosting overall accuracy;
 \item Our method is also the first to incorporate self-consistency based on temporal ensembling directly in a co-training framework. We demonstrate empirically that self-paced learning and self-consistency regularization play a complementary role in semi-supervised segmentation, and that their combination leads to state-of-the-art performance; 
 \item We also present an uncertainty-regularized version of our self-paced JSD loss, which further leverages entropy regularization to enforce joint confidence across the different models. By doing so, confident predictions for unlabeled images are learned better;
 \item We perform one of the most comprehensive analyses of co-training for semi-supervised medical image segmentation, evaluating the proposed method on three different segmentation tasks and against five recently-proposed approaches for this problem. Results demonstrate the advantages of the proposed method compared to existing approaches; 
\end{itemize}
The rest of this paper is structured as follows. In Section \ref{sec:related-works}, we give an overview of relevant literature on semi-supervised segmentation and related work on entropy regularization and self-paced learning. Section \ref{sec:proposed-method} then describes our self-paced and self-consistent co-training method for semi-supervised segmentation. Afterwards, Section \ref{sec:experiments} and \ref{sec:results} present the experimental setup and obtained results. We conclude the paper with a summary of our work's main contributions and its potential extensions.

\begin{figure*}[t!]
\centering
\includegraphics[width=.85\textwidth]{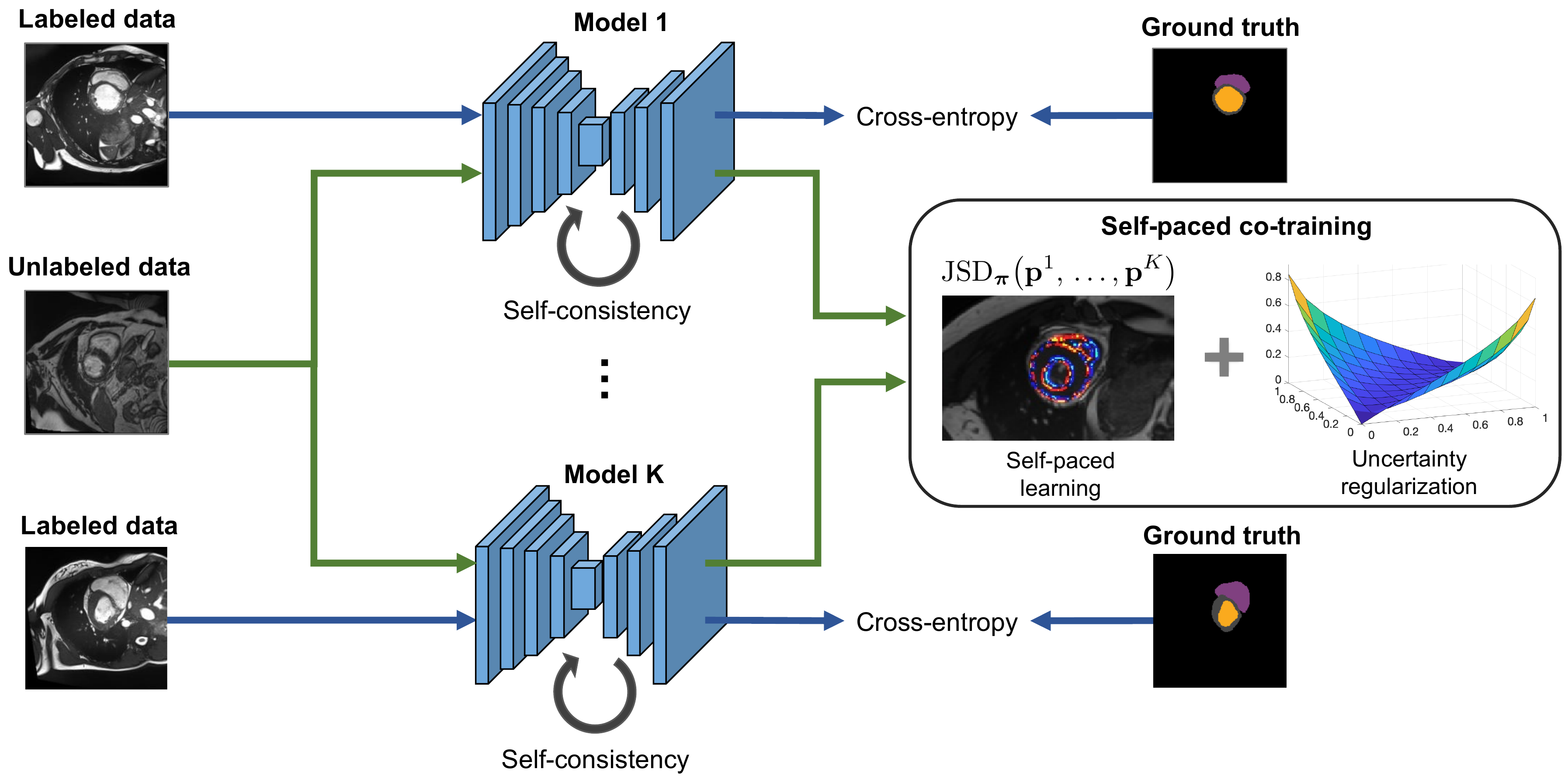}
\caption{\textbf{Diagram of our proposed self-paced and self-consistent co-training for semi-supervised segmentation}. The proposed method consists of three different losses: 1) a pixel-wise supervised loss $\loss_{\text{sup}}$ for labeled images; 2) a self-paced co-training loss $\loss_{\text{spc}}$ encouraging the $K$ segmentation models to agree on increasingly harder regions in unlabeled images; 3) a self-consistency loss $\loss_{\mr{reg}}$ based on temporal ensembling that regularizes the prediction of individual models.}
\label{fig:diagram_method}
\end{figure*}

\section{Related work}\label{sec:related-works}
We start by an overview of recent work on semi-supervised segmentation. We then focus our presentation on entropy regularization techniques for computer vision, and particularly entropy minimization approaches for segmentation. Last, we present recent literature on self-paced learning for image analysis, which is at the core of our proposed method. 

\mypar{Semi-supervised segmentation} The bulk of semi-supervised methods for segmentation can be roughly grouped into four categories: self-training methods~\citep{zou2018unsupervised,zhu2020improving,bai2017semi}, regularization methods~\citep{chaitanya2019semi,zhao2019data,bortsova2019semi,cui2019semi,perone2019unsupervised,yu2019uncertainty,dou2020unpaired}, adversarial learning~\citep{souly2017semi,zhang2017deep,hung2018adversarial,mondal2018few}, and co-training methods~\citep{peng2019deep,xia20203d,zhou2019semi}.

In basic self-training approaches, a model generates pseudo-labels for unlabeled data and is then retained with the updated set of labeled examples. The Pseudo-label algorithm proposed by Lee~\citep{lee2013pseudo} fine-tunes the model with the new pseudo-labeled data instead of retraining it at each pseudo-labeling step. Since pseudo-labels predicted in earlier training stages are generally less reliable, their importance in the loss function is gradually increased over training. When annotated data is very scarce, incorrect predictions may however be reinforced by this approach, leading to a worse performance.

A wide range of regularization-based methods have also been proposed for semi-supervised classification and segmentation. 
The $\Gamma$ model~\citep{rasmus2015semi} evaluates unlabeled data samples with and without noise, and then applies a consistency loss between the two predictions. However, if the model predicts incorrect targets, enforcing consistency on wrong predictions may impede learning. To mitigate this problem, \cite{miyato2019virtual} propose a virtual adversarial training (VAT) regularization method where a divergence-based local smoothness loss is employed to make the model robust to adversarial perturbations of the input. \cite{Laine17Temporal} present a knowledge distillation method called temporal ensembling which encourages consistent network outputs for different evaluations and dropout conditions of the same input. This is achieved by aggregating the predictions of multiple previous network evaluations into an ensemble (the Teacher), and minimizing the $L_2$ distance between predictions of the ensemble and the current model (the Student). A drawback of this approach is that the learned information is incorporated into the training process at a slow pace since each target is updated only once per epoch. To overcome this limitation, \cite{tarvainen2017mean} developed the Mean Teacher method, which averages and compares model weights instead of predictions. In addition, since weight averages involve all layers, not only the last one, the target model can learn better intermediate representations. 

Recently, several semi-supervised segmentation methods were proposed based on knowledge distillation approaches like Mean Teacher~\citep{cui2019semi,perone2019unsupervised,yu2019uncertainty,dou2020unpaired}. \modifR{\cite{yu2019uncertainty} presented an Uncertainty-Aware Mean Teacher (UA-MT) framework where Monte-Carlo dropout is employed to estimate the pixel-wise prediction uncertainty of the teacher, and uncertainty values are used as weights in the consistency loss between the teacher and student outputs (i.e., output consistency for the teacher's confident predictions is given more importance). While our method also exploits confidence to transfer knowledge across models, there are important differences compared to this existing approach. Whereas UA-MT estimates uncertainty in an extrinsic fashion (i.e., with Monte Carlo dropout sampling) and for a single teacher network, the proposed model considers the prediction confidence of multiple networks directly in the objective function. Specifically, our model includes self-paced learning weights which control the reliability of a pseudo-label for a given pixel as separate variables in the learning process, and solves a global optimization problem including both these weights and network parameters. In contrast, UA-MT relies on a manually-selected threshold on the dropout uncertainty to select confident predictions. Another important difference with the work  of \cite{yu2019uncertainty} is that our method also incorporates an uncertainty regularization loss that encourages the trained networks to both agree with each other and be confident in their prediction}. 

Whereas these approaches are agnostic to the segmentation task, \cite{chaitanya2019semi} proposed a data augmentation method where a generative model is trained with task-specific data to generate realistic images and segmentation masks. Similar techniques based on transformation consistency are presented in~\citep{zhao2019data,bortsova2019semi}. In our experiments, we show that the proposed method outperforms state-of-the-art regularization-based approaches for semi-supervised segmentation.

Adversarial learning methods for semi-supervised segmentation typically use a classification network (the discriminator) during training to predict if the output of the segmentation network (the generator) is from the same or different distribution compared to labeled examples~\citep{zhang2017deep,luc2016semantic,souly2017semi,mondal2018few}. By trying to fool the discriminator, the generator is encouraged to output a similar predictive distribution for both images with and without annotations. A potential issue with this approach is that the adversarial network can have a reverse effect, where the output for annotated images becomes increasingly similar to the wrong predictions for unlabeled images. A related strategy proposed by \cite{hung2018adversarial} uses the prediction of a fully-convolutional discriminator at each pixel as a confidence map for the segmentation. For unlabeled images, predictions in high-confidence regions are then considered as pseudo-labels to update the segmentation network. Despite their success, adversarial learning approaches are often avoided due to the complexity and instability of their training.

Co-training methods have also shown promising results for semi-supervised segmentation. \cite{peng2019deep} introduce a deep co-training method which combines a consistency loss based on JSD and a model diversity loss using adversarial training. \cite{zhou2019semi} use different planes of a 3D MRI scan as separate co-training views and use the aggregated prediction on unlabeled images to guide the learning. In their paper, \cite{xia20203d} extend this last framework by considering Bayesian-estimated uncertainty when merging the predictions of different views. While this approach considers prediction uncertainty, it does so without taking into account learning pace. In comparison, our proposed method provides a principled technique for self-paced co-training based on a generalized JSD, where high-confidence predictions are leveraged in a dynamic fashion to co-regularize the segmentation networks. To our knowledge, our work is also the first to propose self-consistency regularization within co-training.

\mypar{Entropy regularization} Entropy minimization was first proposed in~\citep{grandvalet2005semi} to improve learning in semi-supervised classification. The basic idea of this approach is encouraging a model to have confident predictions for unlabeled examples by minimizing their entropy. 
This forces the decision boundary to go through a low-density region of the data, thereby helping the classifier generalize to unseen examples. 
Although this approach has shown great potential for unsupervised domain adaptation~\citep{vu2019advent}, its application to semi-supervised segmentation remains limited. In this work, we extend the concept of semi-supervised entropy regularization, which has until now been used in a single-model scenarios, to the more general multi-view co-training setting.

\mypar{Self-paced learning} Self-paced learning (SPL)~\citep{kumar2010self} extends the paradigm of curriculum learning~\citep{bengio2009curriculum}, where a model is learned by adding gradually more difficult instances during training. The standard SPL model assigns a weight to each instance in the learning objective and adds a self-paced regularizer that determines these weights for a given learning pace. So far, only few works have explored self-paced learning for segmentation. \cite{wang2018deep} present an SPL method for lung nodule segmentation where the weight of each 3D image in the loss is controlled by the SPL regularizer. Recently, \cite{ma2020self} proposed a first self-paced approach for multi-view co-training. Whereas standard co-training techniques adopt a ``draw without replacement strategy'' which may lead to learning incorrect predictions, this approach adds co-regularization terms in the loss function to select pseudo-labeled instances dynamically during training. Our method, which extends the standard JSD agreement loss to consider prediction uncertainty, significantly differs from this approach based on cross-view correlation. Moreover, while the approach in~\citep{ma2020self} requires an alternating optimization scheme, where pseudo-labels are updated separately from network parameters, our self-paced co-training model can be trained in an end-to-end manner without explicitly computing pseudo-labels.

\section{The proposed method}\label{sec:proposed-method}
An overview of the proposed method for semi-supervised segmentation is shown in Fig. \ref{fig:diagram_method}. Let $\lab=\{(\img_s,\gt_s)\}_{s=1}^{|\lab|}$ be a small set of labeled examples, where each $\img_s \in \Real^{|\pixels|}$ is an image and $\gt_s \in \{0,1\}^{|\pixels|\times|\classes|}$ is the corresponding ground-truth segmentation mask. Here, $\pixels \subset \ZZ^2$ denotes the set of image pixels and $\classes$ the set of segmentation classes. Given labeled images $\lab$ and a larger set of unlabeled images $\unlab=\{\img_u\}_{u=1}^{|\unlab|}$, the proposed co-training method learns $K$ segmentation networks corresponding to the different views. Each network $f^k$ is parameterized by a set of weights $\params^k$. 

Labeled and unlabeled images are exploited jointly during training by minimizing the following loss function:
\begin{equation}
\Loss_{\mr{total}} \, = \, \Loss_{\mr{sup}}(\params;\,\lab) \, + \, \lambda_{1}\Loss_{\mr{\textsc{spc}}}(\params;\,\unlab) + \lambda_{2}\Loss_{\mr{reg}}(\params;\,\unlab).
\label{eq:total_loss}
\end{equation}
The supervised loss $\Loss_{\mr{sup}}(\cdot)$ encourages individual networks to predict segmentation outputs for labeled data that are close to the ground truth:
\begin{equation}
\Loss_{\mr{sup}}(\params;\lab) \, = \, \frac{1}{K|\lab|}\sum_{k=1}^K \sum_{(\img_s,\gt_s)\in\lab} \!\!\loss_{\mr{sup}}\big(f^k(\img_s; \,\params^k), \gt_s\big)
\label{eq:sup}
\end{equation} 
While any segmentation loss can be considered, in this work, we use the well-know cross-entropy loss for $\loss_{\mr{sup}}(\cdot)$. Let $\pp = f(\img; \,\params) \in [0,1]^{|\pixels|\times|\classes|}$ be the class probability map predicted by a network $f$, cross-entropy is defined as
\begin{equation}\label{eq:supDtail}
\loss_{\mr{sup}}(\pp, \gt) \, = \, -\sum_{i\in \pixels}\sum_{j\in\classes} y_{ij} \log(p_{ij}).
\end{equation}
Two separate loss terms are employed to leverage unlabeled data in the learning process. The first one, $\Loss_{\mr{\textsc{spc}}}(\cdot)$, implements our self-paced co-training strategy that lets segmentation networks learn from gradually-harder examples over training epochs. This strategy also incorporates an uncertainty regularizer encouraging models to become confident during training. The second loss term, $\Loss_{\mr{reg}}(\cdot)$, enhances learning by applying self-consistency regularization on the models of each view. This regularization technique, based on temporal ensembling, improves the robustness of individual networks when training them with few labels. In the following subsections, we present the proposed self-paced learning and self-consistency regularization co-training losses in greater details.

\subsection{Self-paced co-training}\label{sec:new-formulation}
We propose a self-paced strategy where high-confidence regions of unlabeled images are first considered in the loss, and those with lower confidence are gradually incorporated during training. We define our self-paced co-training task as the following optimization problem:

\begin{align}\label{eq:spct}
& \min \ \LossSPCT(\{\params^k\}, \{\vec{\pslab}_{ui}\}, \{\ww_{ui}\}) \ = \ \nonumber\\ 
&  \ \ \frac{1}{|\unlab|}\sum_{\img_u \in \unlab} \sum_{k=1}^K \sum_{i \in \pixels}
 w_{uik} \, \kl{\pp^k_{ui}}{\vec{\pslab}_{ui}} \, + \, \reg_{\gamma}(w_{uik}) \\ 
 & \tx{s.t.} \ \sum_{j \in \classes} \pslab_{uij} \, = \, 1, \ \forall u,\forall i\nonumber\\
 & \phantom{\tx{s.t.}} \ \pslab_{uij} \in [0,1], \, \forall u, \forall i, \forall j; \
 w_{uik} \in [0,1], \, \forall u,\forall i,\forall k\nonumber
\end{align}

In this formulation, $\vec{\pslab}_{ui}$ is the soft pseudo-label vector of pixel $i$ in unlabeled image $\img_u$, $D_{\mr{KL}}$ is the Kullback–Leibler (KL) divergence which imposes the prediction $\pp^k_{ui}$ of each network $f^k$ to agree with $\vec{\pslab}_{ui}$. The importance of pixel $i$ in $\img_u$ on the loss for model $f^k$ is controlled by self-paced learning weight $w_{uik} \in [0,1]$ that is optimized jointly with $\pp^k_{ui}$ and $\vec{\pslab}_{ui}$. As in traditional self-paced learning methods \citep{kumar2010self}, $\reg$ is a regularization function parameterized by a learning pace parameter $\gamma\!\geq\!0$, such that $w_{uik}$ decreases monotonically with $\kl{\pp^k_u}{\vec{\pslab}_u}$ and increases monotonically with $\gamma$. In other words, $w_{uik}$ should be smaller for pixels $i$ of an image $\img_u$ that are more difficult to predict for model $f^k$ (i.e., having a larger loss) and should increase when using a larger learning pace $\gamma$. 

\subsubsection{Choice of self-paced regularization} 
A common choice for regularization term $\reg$ is a simple linear function: $\reg_{\gamma}(w_{uik}) = -\gamma \, w_{uik}$. This choice leads to a binary solution where $w_{uik} = 1$ if $\kl{\pp^k_u}{\vec{\pslab}_u} \leq \gamma$ and $w_{uik} = 0$ otherwise \citep{kumar2010self}. In our setting, this simple solution has two considerable drawbacks. First, since the weights are binary, all selected pixels (i.e., $w_{uik} = 1$) for a given learning pace $\gamma$ have an equal importance in the loss. This is similar to a hard attention mechanism, which typically performs worse than soft-attention for vision tasks~\citep{fu2019dual}. More importantly, if $\gamma$ is set too small, very few pixels will contribute to the loss, thereby impeding the learning. To alleviate these problems, we instead employ the following quadratic regularization function: %$.
$\reg_{\gamma}(w_{uik}) = \gamma\big(\tfrac{1}{2}w^2_{uik} - w_{uik})$. 

\subsubsection{Self-paced co-training loss and weights update}
As in standard self-paced learning techniques, the learning weights and model parameters are optimized separately in an alternating manner. However, unlike the approach in \citep{kumar2010self}, our method does not require the explicit computation of pseudo-labels. Learning weights are updated according to the following theorem.
\begin{theorem1}
Given fixed model parameters $\{\params^k\}$ and pseudo-labels $\{\vec{\pslab}_{ui}\}$, the optimal learning weights can be obtained as
\begin{equation}\label{eq:learning-weights}
w^*_{uik} \ = \ \max\Big(1 - \frac{1}{\gamma}\kl{\pp^k_{ui}}{\vec{\pslab}_{ui}}, \, 0\Big)
\end{equation}
\end{theorem1}
\vspace{1mm}
The proof of Theorem 1 is shown in \ref{sec:proof1}.

In practice, we replace the $0$ in (\ref{eq:learning-weights}) by a small $\epsilon\!>\!0$ to avoid zero-divison in subsequent steps. We see that the self-paced regularization term acts as a soft attention mechanism where the importance of a pixel on the loss of model $f^k$ is inversely related to the divergence between the prediction of model $f^k$ and the pseudo-label for that pixel. 

We show in the next theorem that the problem of finding pseudo-labels and network parameters amounts to minimizing a generalized form of Jensen-Shannon divergence $\JSD_{\ppi}\big(\pp^1,\, \ldots, \pp^K\big) = \entr\big(\sum_k \pi_k \, \pp^k\big) - \sum_k \pi_k\, \entr\big(\pp^k\big)$.

\begin{theorem1}
For a fixed set of learning weights $\{\ww_{ui}\}$, learning the model parameters $\{\params^k\}$ for optimal pseudo-labels corresponds to the following problem: 
\begin{align}\label{eq:spct-loss}
& \min_{\{\params^k\}} \ \, \frac{1}{|\unlab|} \sum_{\img_u \in \unlab}\sum_{i \in \pixels} \rho_{ui} \, \JSD_{\ppi_{ui}}\big(\pp^1_{ui}, \, \ldots, \pp^K_{ui}\big) \\
& \quad \tx{with } \ \pi_{uik} \, = \, \frac{w_{uik}}{\rho_{ui}}; \quad \rho_{ui} \, = \, \sum_{k=1}^K w_{uik}\nonumber
\end{align}
\end{theorem1}

The proof of Theorem 2 is shown in \ref{sec:proof2}.

Intuitively, the formulation in (\ref{eq:spct-loss}) imposes individual networks to give, for each unlabeled image pixel, a prediction similar to the confidence-weighted average of all models. %The confidence is determined by learning pace $\gamma$. 
Additionally, the importance of each pixel in the total loss is weighted by coefficient $\rho_{ui}$ that corresponds to the total confidence of models for this pixel. 

\subsubsection{Setting the learning pace parameter}\label{sec:learning-pace}
A common challenge in self-paced learning methods is finding a suitable value for the learning pace parameter $\gamma$: an overly small $\gamma$ will select too few pixels for co-training, which impedes learning, whereas a too large $\gamma$ will ignore the relative difficulty of individual pixels and amounts to having no self-paced learning. To find a good range of values for $\gamma$, we consider the optimal solution for the self-paced weights $w_{uik}$ in (\ref{eq:learning-weights}) and for pseudo-labels $\vec{\pslab}_{ui}$ in (\ref{eq:pseudol-def}). Combining both, we have that pixel $i$ of image $\img_u$ is selected for model $f^k$ (i.e., non-zero weight $w_{uik}$) if $\kl{\pp^k_{ui}}{\sum_{k'} \pi_{uik}\,\pp^{k'}_{ui}} \leq \gamma$. As it is a divergence, the left-side term of this inequality has a lower-bound of 0. Following \citep{lin1991divergence}, it can also be shown that this term is upper-bounded by  $-\log_2(\pi_{uik})$. Furthermore, if we use a small $\epsilon$ instead of zero when updating the weights in (\ref{eq:learning-weights}), we have that $w_{uik} \in [\epsilon,1]$ and thus $-\log_2(\epsilon/K) = \log_2(K/\epsilon)$ is also an upper bound. Summing up, $\gamma$ can be increased following the $[0, \log_2(K/\epsilon)]$ range. 

\begin{figure}[t!]
\centering
\renewcommand{\arraystretch}{1}
\setlength{\tabcolsep}{0pt}
\begin{small}
\begin{tabular}{cc}
\alphaJSD{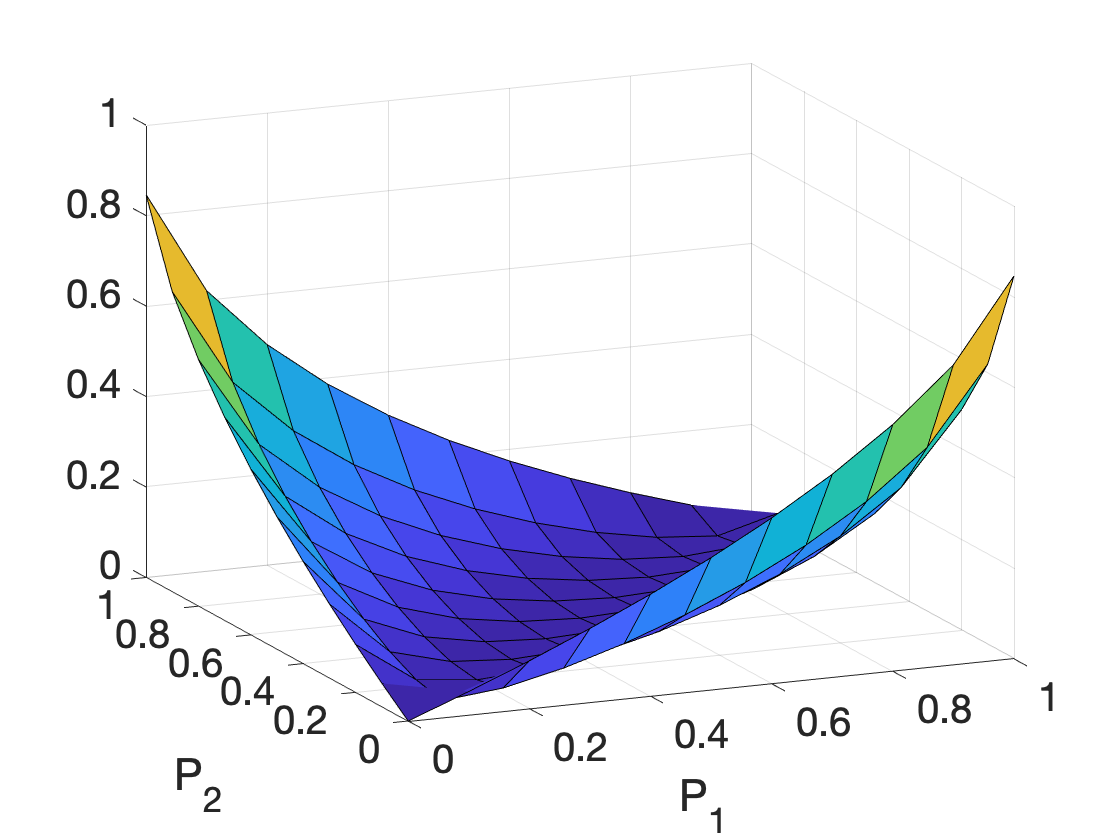} & \alphaJSD{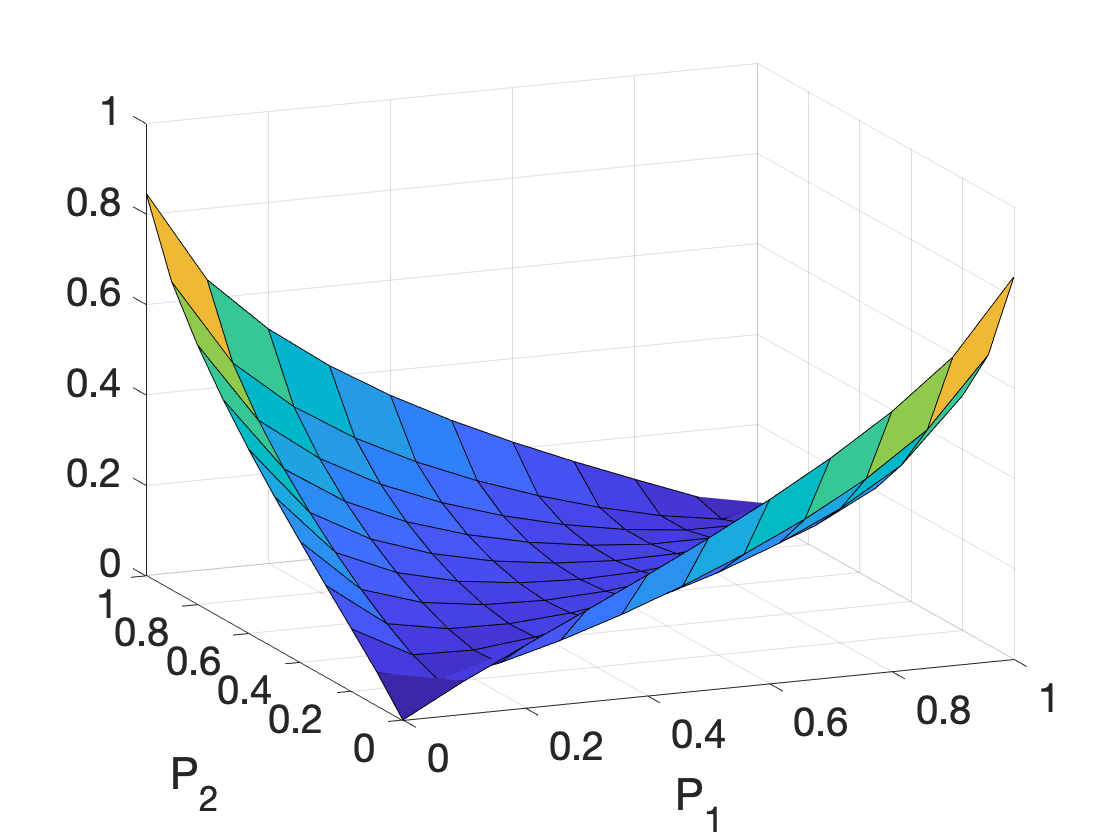} \\
$\alpha = 0.0$ & $\alpha = 0.1$ \\[-2pt]
\alphaJSD{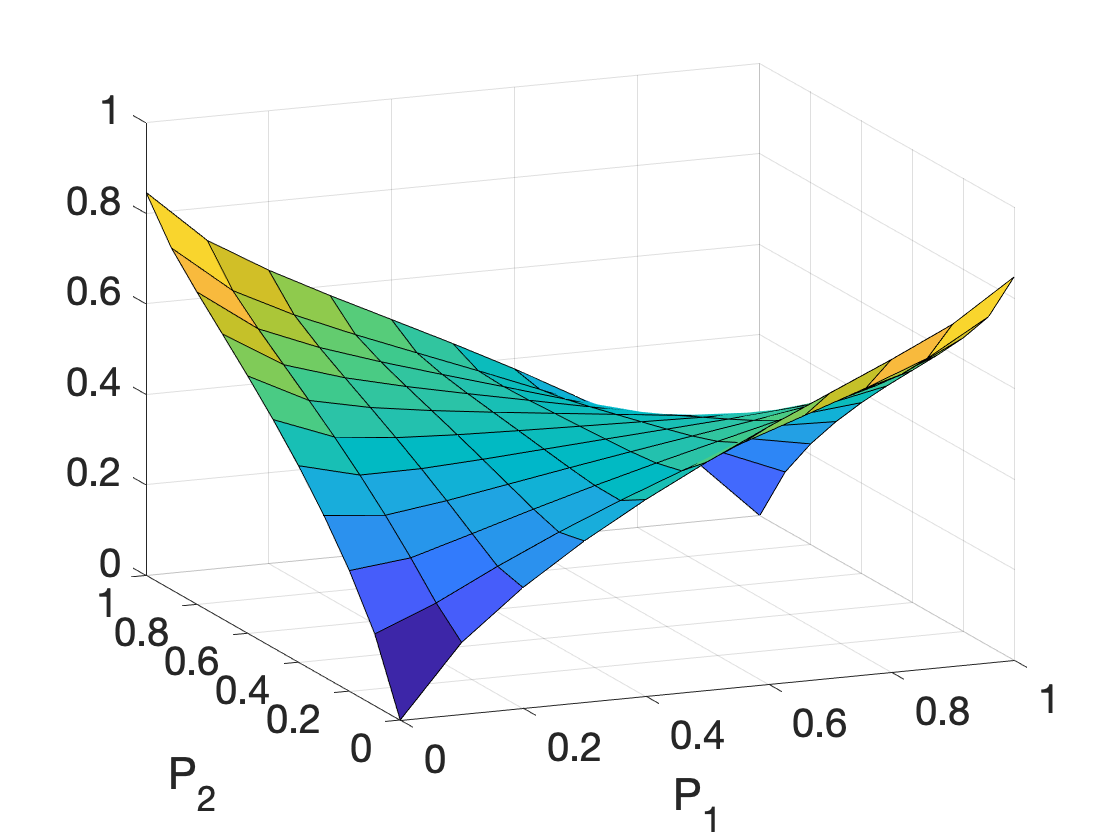} & \alphaJSD{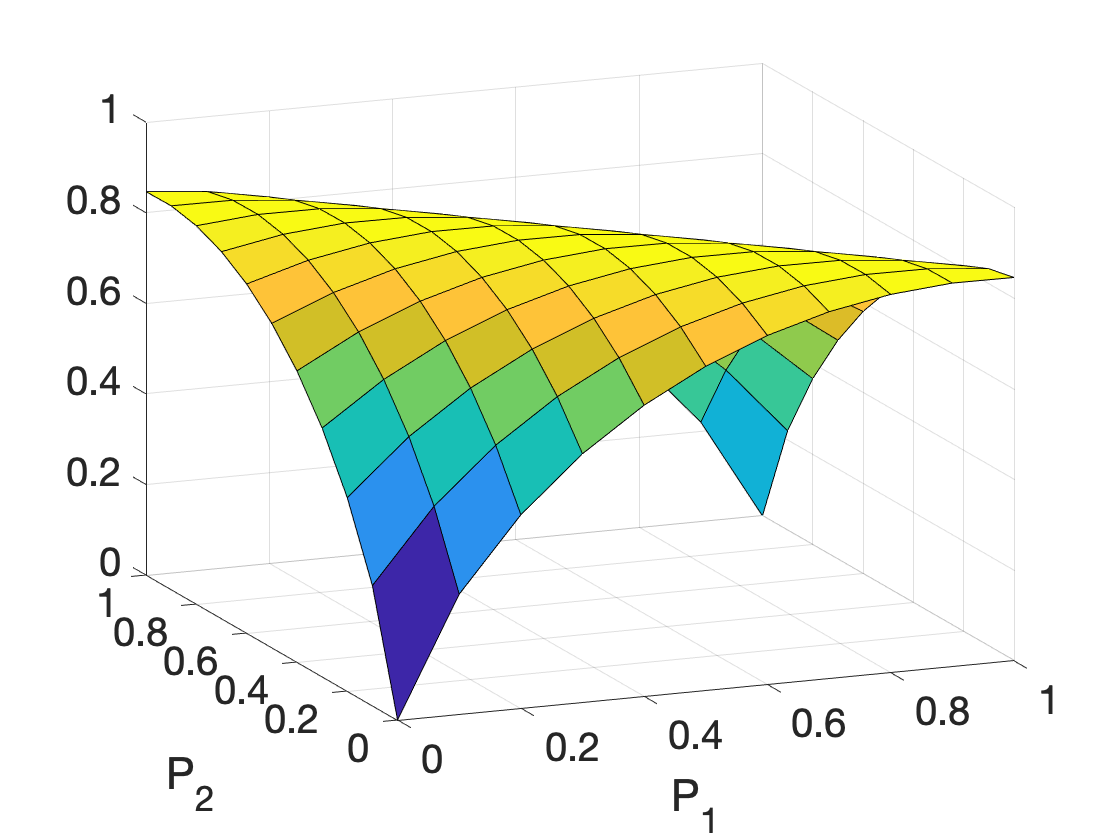} \\
$\alpha = 0.5$ & $\alpha = 1.0$ \\
\end{tabular}
\end{small}
\caption{Illustration of the proposed entropy regularized JSD between two Bernoulli distributions $P_1$ and $P_2$, for different $\alpha$ values. When using $\alpha$\,=\,0, we have the standard JSD which is zero when $P_1$\,=\,$P_2$ regardless of the confidence (i.e., entropy). As $\alpha$ is increased toward 1, the loss encourages both the agreement and confidence of distributions.}
\label{fig:alpha_motivation}
\end{figure}

\subsection{Uncertainty regularization}\label{sec:uncertainty-reg}
As defined in (\ref{eq:spct-loss}), our self-paced co-training method, based on generalized JSD, extends the traditional multi-view learning approach where inter-model agreement is typically measured with the standard JSD. An important drawback of JSD is that it enforces models to agree with each other, but does not require them to be confident in their prediction. This is illustrated in Fig.~\ref{fig:alpha_motivation} (top left), where we show the JSD between two Bernoulli distributions $P_1$ and $P_2$ (i.e., $P$ is the probability of class 1, and $1\!-\!P$ the probability of class 2). As can be seen, a JSD of 0 is obtained when $P_1 = P_2$, whether the predictions are confident (e.g., $P_2 = P_2 = 1$) or not (e.g., $P_1 = P_2 = 0.5$). 

A powerful technique for semi-supervised learning, called entropy minimization~\citep{grandvalet2005semi}, consists in increasing the confidence of predictions for unlabeled examples. Based on this idea, we propose an entropy regularizer over JSD that encourages models to agree while also making them confident. This regularized divergence, which is parameterized by $\alpha \in [0,1]$, is defined as follows:
\begin{equation}\label{eq:jsdAlpha}
\mr{JSD}^{\alpha}\big(\pp^1, \ldots, \pp^K) \, = \, \entr\Big(\frac{1}{K}\sum_{k=1}^K\pp^k\Big) \, - \, \frac{(1\!-\!\alpha)}{K}\sum_{k=1}^K\entr\big(\pp^k\big).
\end{equation}
When $\alpha\!=\!0$, we get the standard definition of JSD. On the other hand, for $\alpha\!=\!1$, $\JSD^{\alpha}$ simply measures the entropy of the mean prediction which is 0 only when 1) all models are 100\% confident and 2) the models agree with each other. Since entropy is non-negative, we have that $\mr{JSD}^\alpha \geq \mr{JSD}^{\alpha'}$ for any $\alpha \geq \alpha'$. As shown in Fig.~\ref{fig:alpha_motivation}, while using $\alpha\!>\!0$ increases the confidence of predictions, it also makes the function non-convex (standard JSD is convex). This may pose a problem in initial stages of training, since incorrect predictions are forced to become more confident until the optimization gets stuck in a poor local minimum. In practice, we avoid this issue by using the convex JSD at the beginning of training and then increasing $\alpha$ slowly to reach a fixed value.

Incorporating the proposed entropy regularization in our self-paced co-training method, we finally get the following loss:
\begin{equation}\label{eq:loss_spc_reg}
\Loss_{\mr{\textsc{spc}}}(\params;\unlab) \ = \
 \frac{1}{|\unlab|} \sum_{\img_u \in \unlab}\sum_{i \in \pixels} \rho_{ui} \, \JSD^{\alpha}_{\ppi_{ui}}\big(\pp^1_{ui}, \, \ldots, \pp^K_{ui}\big)
\end{equation} 
where $\rho_{ui}$ and $\ppi_{ui}$ are defined as in (\ref{eq:spct-loss}), and
\begin{equation}
\JSD^{\alpha}_{\ppi}\big(\pp^1, \, \ldots, \pp^K\big) \, = \, \entr\big(\sum_{k=1}^K \pi_k \, \pp^k\big) \, - \, \big(1-\alpha\big)\!\sum_{k=1}^K\!\pi_k\, \entr\big(\pp^k\big).
\end{equation}

\subsection{Self-consistent co-training}\label{sec:self-consistent}
While co-training promotes consistency across different models, it has also been shown that imposing consistency between the predictions of a single model at different training iterations can also improve its robustness in a semi-supervised setting \citep{tarvainen2017mean}. Based on this idea, we propose to incorporate a self-ensembling strategy in our co-training method, where we replace each segmentation model $f^{k}$ by two models: a Teacher $f^{k}_{T}$ and a Student model $f^{k}_S$. The two models share the same architecture, however the Teacher's parameters are a temporal ensembling of the student's across different training steps. Specifically, we update the Teacher's parameters at step $t$, denoted as $\params^{(t)}_T$, using an exponential moving average of the Student's parameters $\params^{(t)}_S$:
\begin{equation}
\label{eq:ema}
\params^{(t)}_{T} \, = \, \beta\params^{(t-1)}_{T} \, + \, (1 - \beta)\params^{(t)}_S.
\end{equation}
The Teacher closely follows the Student for $\beta\!\approx\!0$, whereas it has a longer-term memory when $\beta$ is near 1. In the latter case, each step the Student takes contributes slightly to the Teacher and, therefore, the Teacher is likely to have a smoother learning. Following the literature~\citep{tarvainen2017mean}, we adopt a $\beta$ of 0.99.

We impose a self-consistency loss that minimizes the $L_2$ distance between the predictions of Teacher-Student pairs for unlabeled images $\img_u \in \unlab$ under random geometric transformations $\tau$:    
\begin{equation}
\Loss_{\mr{reg}}(\params; \unlab) \, = \, \frac{1}{K\,|\unlab|}\sum_{k=1}^{K}\sum_{\img_u\in \unlab}\big\|\tau\big(f^{k}_{T}(\img_u)\big) - f^{k}_{S}\big(\tau(\img_u)\big)\big\|^2.
\end{equation}
Here, the same transformation $\tau$ is applied to the Teacher's prediction so that it aligns with the Student's output for the transformed input image. In this work, we considered random rotations for $\tau$. 
Moreover, while KL divergence can also be employed to measure prediction agreement, like most temporal-ensembling approaches, we used $L_2$ distance as it leads to a smoother optimization. Unlike $L_2$ which has a bounded gradient, the gradient of KL may become large, especially in initial training iterations. Thus, it can overpower the gradient of the supervised loss and cause the learning to fail. The overall optimization process, combining all three loss terms, is summarized in Algorithm \ref{algo}. 

{\centering
\SetAlFnt{\small}

\begin{algorithm2e}[t!]

\SetNoFillComment

\KwIn{Labeled dataset $\lab$, unlabeled dataset $\unlab$, and number of co-training models $K$;}
\KwOut{Model parameters $\{\params^k\}$;}
\caption{Training of the self-paced and self-consistent co-training model.
}\label{algo}

    \BlankLine
    Randomly initialize network parameters $\params^k$, $\forall k$\;
    Initialize learning pace: $\gamma \, \gets \, \gamma_0$\;
    \BlankLine
    \For{$\mr{epoch} = 1,\,\ldots,\,n_{\mr{epochs}}$}{ %\Do
    \For{$\mr{n} = 1,\,\ldots,\,n_{\mr{iter}}$}{ %\Do
        Sample training batch $\{\lab_n, \unlab_n\}$\; 
        
        For all $\img_u \in \unlab_n$, compute learning weights $w_{uik}$ using (\ref{eq:learning-weights})\; 
        According to (\ref{eq:total_loss}), do a batch gradient descent step on the Student models' parameters $\params^k_{S}$\;
        Update the Teacher models' parameters $\params^k_T$ based on (\ref{eq:ema})\; 
    }
    Adjust the SGD learning rate\;
    Update learning pace: $\gamma \ \gets \ \gamma \times
    \mr{increaseFactor}$\;
    }
\Return{$\{\params^k\}$} \;
\end{algorithm2e}
}

\section{Experiments}\label{sec:experiments}

To evaluate the performance of the proposed self-paced and self-consistent co-training method, we test it on three medical image segmentation tasks involving MRI and CT data. In the next subsections, we describe the datasets and performance metrics employed in the evaluation, the experimental setup, and implementation details of our method.

\subsection{Datasets and metrics}
\subsubsection{Dataset}
Our experiments are conducted on three clinically-relevant benchmark datasets: Automated Cardiac Diagnosis Challenge (ACDC)~\citep{bernard2018deep}, Spleen sub-task dataset of the Medical Segmentation Decathlon Challenge~\citep{simpson2019large}, and the Prostate MR Image Segmentation (PROMISE) 2012 Challenge dataset~\citep{bProstate}.

\mypar{ACDC} This dataset consists of 200 MRI scans from 100 patients, including 20 healthy patients, 20 patients with previous myocardial infarction, 20 patients with dilated cardiomyopathy, 20 patients with an hypertrophic cardiomyopathy, and 20 patients with abnormal right ventricle. Scans correspond to end-diastolic (ED) and end-systolic (ES) phases, and were acquired on 1.5T and 3T systems with resolutions ranging from 0.70$\times$0.70\,mm to 1.92$\times$1.92\,mm in-plane and 5\,mm to 10\,mm through-plane. Three cardiac regions are labeled in the ground-truth: left ventricle (LV), right ventricle (RV) and myocardium (Myo). In our experiments, we treat the extraction of each region as a separate binary segmentation task. For our experiments, we used a split of 75 subjects (150 scans) for training and 25 subjects (50 scans) for testing. Slices within 3D-MRI scans were considered as 2D images, themselves randomly cropped into patches of size 192$\times$192. These patches are fed as input to the network.

\mypar{Spleen} This dataset consists of total 61 CT scans (only 41 were given with ground truth). For our experiments, 2D images are obtained by slicing the high-resolution CT volumes, followed by a max-min normalization to the $[0,1]$ range. Each slice is then resized to 256$\times$256. We split the dataset into training and testing sets, comprised respectively of 36 and 5 CT scans.

\mypar{Prostate} This third dataset, which is provided by Radboud University, consists of 50 3D T2-weighted MRIs of the prostate region with expert annotations for two structures: peripheral zone and central gland. This dataset was split into training and testing sets containing the MRIs of 40 and 10 patients, respectively. Once again, slices within MRIs were treated as 2D images that are randomly cropped into input patches of size 192$\times$192.

\subsubsection{Metrics}
Two metrics are used to evaluate the segmentation accuracy of tested methods: Dice similarity coefficient (DSC) and Hausdorff distance (HD). DSC measures the degree of overlap between the segmentation region and ground truth, and is defined as
\begin{equation}
 \mr{DSC}(G,S) \, = \, \frac{2|S\cap G|}{\,|S|+|G|},
\end{equation}
where $S$ is the predicted labels and $G$ is the corresponding ground truth labels. DSC values range from 0 to 1, a higher value representing a better segmentation.

Haussdorf distance (HD) is a boundary distance metric which measures the largest distance between a point in $S$ and its nearest point in $G$. A smaller HD value indicates a better segmentation. HD is defined as
\begin{equation}
 \mr{HD}(G,S) \, = \, \max\big\{d(G, S), \,d(S, G)\big\}
\end{equation}
where $d(S,G)$ is the maximum of nearest-neighbor distances from $S$ to $G$. In our results, we report the HD in millimeters.

\subsection{Experimental setup}
We first use different levels of supervision to assess the stability of our proposed method. For ACDC and Prostate, we tested the following two settings: $10\%$ and $5\%$ of the training set as labeled data. Since the Spleen dataset is more challenging to segment, considering only $5\%$ of training examples as labeled leads to the collapse of tested algorithms. To avoid this problem, we instead used labeled data ratios of $10\%$ and $7\%$ for this dataset. Moreover, to measure the impact of our self-paced learning and self-consistency losses, we performed an ablation study where we disable one of these losses while keeping the other. Finally, to investigate our method's performance and scalability in a multi-view setting, we tested it with more than two segmentation models. 

We compare our method against several baselines. To have upper and lower bounds on performance, we first include full-supervision and semi-supervision baselines. The full-supervision baseline considers all training examples as labeled. On the other hand, the semi-supervision baseline uses only the partial subset of labeled examples (i.e., 10\%, 7\% or 5\%) and ignores the remaining unlabeled images during training. As our proposed method extends standard co-training, we consider the model with only JSD as another strong baseline called Co-training. Moreover, we compare our method with four state-of-the-art semi-supervised segmentation approaches: Entropy minimization~\citep{grandvalet2005semi}, deep adversarial networks (DAN)~\citep{zhang2017deep}, Mean Teacher (MT)~\citep{tarvainen2017mean}, and uncertainty-aware Mean Teacher (UA-MT)~\citep{yu2019uncertainty}. We keep the same underlying architecture, optimization procedure, and data augmentation across all tested methods. For main experiments, we adopt a dual-view setting for standard co-training and our method. As in~\citep{peng2019deep}, we use soft-voting to aggregate the prediction of individual models into the final segmentation.

To have a fair comparison, we selected the hyper-parameters of all methods based on grid search. For Mean Teacher, which involves two models and a data transformation procedure, we applied data transformations on input images for the Student model and on output predictions for the Teacher. We update the Teacher using (\ref{eq:ema}) and, following \citep{peng2019deep}, use this model's output as the segmentation prediction to measure performance. 

\begin{table*}[!t]
\centering
\caption{Mean DSC (\%) of tested methods on the ACDC dataset, for different ratios of labeled examples in the training set. Bold font values indicate the best performing method for each labeled data setting. Values are underlined if the improvement over all other approaches is statistically significant ($\mr{p} <$ 0.05)}.
\label{tab:ACDC_dice}
\setlength{\tabcolsep}{6pt}
\renewcommand{\arraystretch}{1}
\begin{footnotesize}
\begin{tabular}{cp{65pt}cccccc}
\toprule
\multirow[b]{2}{*}{\B Labeled\,\%} & \multirow[b]{2}{*}{\B Method} & \multicolumn{4}{c}{\B ACDC} \\
\cmidrule(l{6pt}r{6pt}){3-6}
 & & \B RV & \B Myo & \B LV & \B Mean \\ 
\midrule
\multirow{1}{*}{100\,\%} & Baseline & 89.29 \,(0.37) & 88.30 \,(0.15) & 94.10 \,(0.32) & 90.56 \,(0.28) \\ 
\midrule
\multirow{6}{*}{10\,\%} & Baseline & 
77.51 \,(0.87) & 81.56 \,(0.46) & 91.72 \,(0.24) & 83.60 \,(0.53) \\
& Entropy min & 
78.19 \,(2.29) & 82.04 \,(0.40) & 91.84 \,(0.27) & 84.02 \,(0.99) \\ 
& DAN & 
82.81 \,(0.15) & 83.77 \,(0.09) & 91.90 \,(0.25) & 86.16 \,(0.16) \\
& MT & 
82.91 \,(1.55) & 85.35 \,(0.57) & 92.37 \,(0.32) & 86.88 \,(0.81) \\ 
&UA-MT & 
83.63 \,(0.89) & 85.78 \,(0.16) & 92.13 \,(0.56) &  87.18 \,(0.54) \\ 
& Co-training & 80.88 \,(0.53) & 83.69 \,(0.32) & 92.84 \,(0.31) & 85.80 \,(0.39) \\ 
& Ours & \B 83.85 \,(0.51) & \B \signif{86.42} \,(0.29) & \B \signif{93.06} (\,0.07) & \B \signif{87.78} \,(0.29) \\ 
\midrule
\multirow{6}{*}{5\,\%} & Baseline & 72.32 \,(2.47) & 75.69 \,(0.69) & 86.87 \,(0.79) & 78.29 \,(1.31) \\ 
& Entropy min & 73.56 \,(0.51) & 75.47 \,(0.84) & 87.25 \,(1.19) & 78.76 \,(0.85) \\ 
& DAN & 
81.54 \,(0.53) & 80.13 \,(0.45) & 90.96 \,(0.33) & 84.21 \,(0.44) \\
& MT & 79.52 \,(0.98) & 83.08 \,(0.60) & 91.38 \,(0.14) & 84.66 \,(0.57) \\ 
& UA-MT & 
81.71 \,(0.62) & 83.08 \,(0.23) & 90.69 \,(0.83) & 85.16 \,(0.56) \\ 
& Co-training & 75.37 \,(1.35) & 79.41 \,(0.41) & 90.72 \,(0.05) & 81.83 \,(0.60) \\ 
& Ours & \B \signif{82.33} \,(0.16) & \B \signif{84.46} \,(0.22) & \B \signif{92.47} \,(0.14) & \B \signif{86.42} \,(0.17) \\ 
\bottomrule
\end{tabular}
\end{footnotesize}
\end{table*}

\begin{table*}[!t]
\centering
\caption{Mean Hausdorff distance (HD) of tested methods on the ACDC dataset. Bold font values indicate the best performing method for each labeled data setting.}
\label{tab:ACDC_HD}
\setlength{\tabcolsep}{6pt}
\renewcommand{\arraystretch}{1}
\begin{footnotesize}
\begin{tabular}{cp{65pt}rrrr}
\toprule
\multirow[b]{2}{*}{\B Labeled\,\%} & \multirow[b]{2}{*}{\B Method} & \multicolumn{4}{c}{\B ACDC} \\ 
\cmidrule(l{6pt}r{6pt}){3-6}
 & & \multicolumn{1}{c}{\B RV} & \multicolumn{1}{c}{\B Myo} & \multicolumn{1}{c}{\B LV} & \multicolumn{1}{c}{\B Mean} \\ 
\midrule
\multirow{1}{*}{100\,\%} & Baseline & 6.05 \,(0.49) & 3.99 \,(0.07) & 2.87 \,(0.08) & 4.30 \,(0.21) \\ 
\midrule
\multirow{6}{*}{10\,\%} & Baseline & 15.10 \,(2.13) & 11.00 \,(2.84) & 6.20 \,(1.29) & 10.77 \,(2.09) \\
& Entropy min & 16.85 \,(2.12) & 8.03 \,(0.48) & 5.77 \,(0.82) & 10.22 \,(1.41) \\ 
& DAN & 
14.00 \,(0.11) & 6.28 \,(0.82)
& 4.46 \,(1.21) & 8.25 \,(0.71) \\
& MT & 12.30 \,(0.02) & 6.97 \,(1.00) & 4.89 \,(1.11) & 8.05 \,(0.71) \\ 
& UA-MT & 
17.12 \,(6.36) & 8.02 \,(0.45) & 6.92 \,(0.38) & 10.69 \,(2.40) \\ 
& Co-training & \B 9.17 \,(0.60) & 6.61 \,(0.35) & 5.01 \,(0.49) & 6.93 \,(0.48) \\ 
& Ours & 10.10 \,(0.68) & \B \signif{5.43} \,(0.40) & \B 3.81 \,(0.58) & \B \signif{6.45} \,(0.56) \\ 
\midrule
\multirow{6}{*}{5\,\%} & Baseline & 23.74 \,(5.97) & 17.27 \,(3.39) & 11.48 \,(3.78) & 17.50 \,(4.38) \\ 
& Entropy min & 24.45 \,(1.50) & 17.30 \,(2.76) & 12.89 \,(3.60) & 18.21 \,(2.62) \\ 
& DAN & 
12.38 \,(0.43) & 8.14 \,(1.32) & 7.04 \,(1.54) & 9.19 \,(1.10) \\
& MT & 14.13 \,(1.93) & 12.55 \,(7.28) & 11.02 \,(1.65) & 12.57 \,(3.62) \\
& UA-MT & 
14.85 \,(1.27) & 11.98 \,(1.94) & 10.06 \,(2.35) & 12.30 \,(1.85) \\ 
& Co-training & 11.07 \,(0.83) & 7.00 \,(0.24) & 5.25 \,(0.11) & 7.77 \,(0.39) \\ 
& Ours & \B 10.22 \,(0.43) & \B \signif{6.30} \,(0.49) & \B 4.93 \,(0.46) & \B \signif{7.15} \,(0.46) \\ 
\bottomrule
\end{tabular}
\end{footnotesize}
\end{table*}

\subsection{Implementation details}
Although any 2D segmentation network can be used, we employed the popular light-weight architecture ENet~\citep{Paszke2016} as the underlying segmentation network, as it offers a good trade-off between accuracy and speed. This architecture, which was developed for efficient semantic segmentation, contains about 84 times less trainable parameters than the well-known U-Net architecture (0.37\,M compared to 31.04\,M for U-Net). It employs a convolutional block with short skip connections, called bottleneck block, and is comprised of different 7 stages: an initial stage of regular convolutions, followed by 5 stages with different numbers of bottleneck blocks, and a final stage of 1$\times$1 convolutions to generate the final segmentation probability map. 

All experiments were carried out using the same training setting with the ENet architecture, rectified Adam optimizer and learning rate warm-up strategy based on cosine decay~\citep{loshchilov2016sgdr}. \modifR{Training was performed for a total of 100 epochs, each one including 200 update iterations. The learning rate was initialized to $1\times10^{-3}$ for ACDC, $1\times10^{-4}$ for Prostate and $3\times10^{-5}$ for Spleen. For all datasets, this learning rate was increased by a factor of 300 in the first 10 epochs and then decreased with a cosine scheduler for the following 90 epochs.}

We used standard data augmentation techniques on-the-fly to avoid over-fitting, including randomly cropping and rotation. For a fair comparison, we keep data augmentation the same across different methods for each segmentation task. For our proposed method, the total loss consists of three terms: supervision loss $\Loss_{\mr{sup}}$ for labeled data, self-paced co-training loss $\Loss_{\mr{\textsc{spc}}}$ and self-consistency loss $\Loss_{\mr{reg}}$ for unlabeled data. As defined in (\ref{eq:total_loss}), these loss terms are balanced with two hyper-parameters $\lambda_1$ and $\lambda_2$. \modifR{Based on grid search, these coefficients were set as follows: $\lambda_{1}\!=\!0.5$, $\lambda_{2}\!=\!4$  for ACDC, $\lambda_{1}\!=\!0.1$, $\lambda_{2}\!=\!4$  for Prostate, and $\lambda_{1}\!=\!0.5$, $\lambda_{2}\!=\!12$ for Spleen. For all three datasets, $\alpha$ was initialized to 0 and then gradually increased to a value near $\alpha\!=\!10^{-4}$ over training. Using a larger value for $\alpha$ led to a worse performance. We believe this is due to the fact that entropy is a concave function with exponentially growing gradient near values of 0 and 1. A small $\alpha\!=\!10^{-4}$ is therefore necessary to avoid the gradient from this term dominating the learning. A small coefficient was also used in previous approaches based on entropy regularization~\citep{grandvalet2005semi}.}

\modifR{The learning pace parameter $\gamma$ was set based on the strategy presented in Section \ref{sec:learning-pace}.
%As mentioned in Section \ref{sec:new-formulation}, the value of self-paced learning weights should monotonically decreasing with respect to the KL divergence, and monotonically increasing with respect to the learning pace parameter. 
%For dataset with different levels of difficulty in segmentation, models' sensitivity to difficulty examples is vary. In order to guarantee the learning efficiency, setting different initial learning pace for different dataset is necessary. In our experiments, after many trials, 
To ensure that the prediction for some pixels is used in the loss of Eq. (\ref{eq:spct-loss}), i.e., that not all learning weights are $0$, we set the initial pace $\gamma_0$ as follows: $\gamma_0\!=\!0.2$ for ACDC and Spleen with 10\% labeled data, $\gamma_0\!=\!0.4$ for Prostate with 10\% and Spleen datasets with 7\% labeled data, and $\gamma_0\!=\!0.5$ for Prostate datasets with 5\% labeled data. In all cases, $\gamma$ was updated so that its upper bound is reached after 50 epochs (see Section \ref{sec:learning-pace} for details).} For all experiments except the multi-view analysis, we trained our method and standard co-training using $K\!=\!2$ segmentation networks. In the multi-view analysis, we compare this setting with using $K\!=\!3$ views. The same hyper-parameters as in previous experiments are used in both cases.

To have a fair comparison, the same grid search strategy was used to select the hyper-parameters of all tested approaches. For each method, we report results obtained for the best combination of hyper-parameters found during grid search. 

\begin{table}[!t]
\centering
\caption{Mean DSC and HD of tested methods on the Prostate dataset, for different ratios of labeled examples in the training set.}
\label{tab:prostate}
\setlength{\tabcolsep}{4pt}
\renewcommand{\arraystretch}{1}
\begin{footnotesize}
\begin{tabular}{cp{50pt}rr}
\toprule
\multirow[b]{2}{*}{\B Labeled\,\%} & \multirow[b]{2}{*}{\B Method} & \multicolumn{2}{c}{\B Prostate}\\
\cmidrule(l{6pt}r{6pt}){3-4}
 & & \multicolumn{1}{c}{\B DSC (\%)} & \multicolumn{1}{c}{\B HD (mm)} \\
\midrule
\multirow{1}{*}{100\,\%} & Baseline & 87.99 (0.20) & 5.04 (0.42) \\
\midrule
\multirow{5}{*}{10\,\%} & Baseline & 63.77 \,(1.76) & 9.55 \,(0.80) \\
& Entropy min & 65.30 \,(3.43) & 11.11 \,(4.57)\\
& DAN & 72.62 \,(1.53) & 10.73 \,(1.44)\\
& MT & 75.27 \,(0.72) & 9.92 \,(1.20) \\
& UA-MT & 75.92 \,(2.77) & 7.22 \,(0.23) \\
& Co-training & 65.09 \,(0.67) & 7.83 \,(0.32) \\
& Ours & \B \signif{76.16} \,(0.62) & \B \signif{6.32} \,(0.54) \\
\midrule
\multirow{5}{*}{5\,\%} & Baseline & 49.97 \,(0.83) & 11.65 \,(5.25) \\
& Entropy min & 50.21 \,(1.93) & 13.52 \,(3.95) \\
& DAN & 61.17 \,(2.43) & 25.51 \,(4.55)\\
& MT & 67.97 \,(1.88) & 11.72 \,(0.76) \\
& UA-MT & 68.91 \,(1.76) & 11.09 \,(0.45) \\
& Co-training & 48.92 \,(2.74) & 9.72 \,(0.92) \\
& Ours & \B \signif{70.15} \,(0.27) & \B \signif{8.18} \,(0.63) \\
\bottomrule
\end{tabular}
\end{footnotesize}
\end{table}

\begin{table}[!t]
\centering
\caption{Mean DSC and HD of tested methods on the Spleen dataset, for different ratios of labeled examples in the training set.}
\label{tab:spleen}
\setlength{\tabcolsep}{4pt}
\renewcommand{\arraystretch}{1}
\begin{footnotesize}
\begin{tabular}{cp{50pt}rr}
\toprule
\multirow[b]{2}{*}{\B Labeled\,\%} & \multirow[b]{2}{*}{\B Method} & \multicolumn{2}{c}{\B Spleen}\\
\cmidrule(l{6pt}r{6pt}){3-4}
 & & \multicolumn{1}{c}{\B DSC (\%)} & \multicolumn{1}{c}{\B HD (mm)} \\
\midrule
\multirow{1}{*}{100\,\%} & Baseline & 94.02 \phm(0.41) & 9.40 \phm\,(4.08) \\
\midrule
\multirow{5}{*}{10\,\%} & Baseline & 61.67 \phm\,(4.15) & 84.08 \phm\,(4.84) \\
& Entropy min & 62.50 \phm\,(1.13) & 71.95 \phm\,(6.89) \\
& DAN & 71.64 \phm\,(1.15) & 119.43 \,(9.71)\\
& MT & 86.95 \phm\,(0.95) & 74.45 \,(13.33) \\
& UA-MT & 85.53 \phm\,(1.44) & 59.88 \phm\,(3.63)\\
& Co-training & 58.95 \phm\,(2.86) & 73.43 \phm\,(7.25)\\
& Ours & \B \signif{88.10} \phm\,(1.57) & \B \signif{30.52} \phm\,(5.23)\\
\midrule
\multirow{5}{*}{7\,\%} & Baseline & 58.84 \phm\,(3.36) & 98.92 \phm\,(2.43) \\
& Entropy min & 61.29 \phm\,(0.10) & 103.09 \phm\,(4.01) \\
& DAN & 64.84 \phm\,(3.02) & 107.54 \,(14.43)\\
& MT & 81.32 \phm\,(3.01) & 65.53 \,(43.63) \\
& UA-MT & 83.37 \phm\,(4.51) & 66.54 \phm\,(3.09) \\
& Co-training & 58.26 \phm\,(0.65) & 85.55 \phm\,(8.45) \\
& Ours & \B \signif{85.50} \phm\,(3.57) & \B \signif{58.77} \phm\,(7.51) \\
\bottomrule
\end{tabular}
\end{footnotesize}
\end{table}

\section{Results}\label{sec:results}

\subsection{Comparison to the state-of-art}
We first compare our self-paced and self-consistent co-training method against the baselines and state-of-the-art approaches for semi-supervised segmentation (i.e., Entropy minimization, Deep adversarial networks, Mean Teacher, Uncertainty-aware Mean Teacher, and Co-training). 

Tables \ref{tab:ACDC_dice} and \ref{tab:ACDC_HD} give the mean DSC and HD obtained by the tested methods for the ACDC dataset. Reported values are averages (standard deviation in parentheses) over 3 runs with different random seeds. As can be seen, all semi-supervised methods yield improvements compared to training without unlabeled images (i.e., partial supervision). Entropy minimization improves the mean segmentation performance by about 0.4\% in terms of DSC compared to the partial supervision baseline, in both the 10\% and 5\% labeled data settings. This confirms the benefit of making the prediction of a single model more confident. Likewise, standard Co-training outperforms partial supervision by 2.2\% for 10\% of labeled data, showing that the collaborative training of two models can improve their individual performance. Mean Teacher, which implements temporal ensembling, achieves a mean DSC gain of 3.3\% for the same labeled data setting. UA-MT further improves this score by 0.3\%, demonstrating the usefulness considering uncertainty in the Mean Teacher framework. However, the best performance on ACDC is obtained by our method with a mean DSC of 87.78\% and 86.42\%, respectively for 10\% and 5\% labeled examples. This DSC performance is only 2.8\% and 4.1\% less than full-supervision. Considering segmentation classes separately, we observe that the highest improvements by our method are for RV (+6.3\%) and Myo (+4.9\%), which are the most difficult regions to segment. The proposed method also leads to an important reduction of HD for all classes and labeled data ratios. Using 5\% of labeled data, our method decreases the mean HD substantially from 17.5\,mm to 7.2\,mm compared to the partial supervision baseline. In comparison, the strong UA-MT baseline yields a much higher mean HD of 12.30\,mm for the same setting. This shows our method's greater ability to regularize segmentation boundaries and avoid large gaps to the ground-truth region.

The performance of our self-paced and self-consistent co-training method is further demonstrated for the Prostate and Spleen datasets in Tables \ref{tab:prostate} and \ref{tab:spleen}. For Prostate, our method achieves significant DSC improvements of 1.2\% when using 5\% of labeled data, compared to the second-best approach (i.e., UA-MT). An even greater performance boost is obtained on the challenging Spleen dataset, with DSC gains of 2.6\% and 2.1\% compared to UA-MT for labeled data ratios of 10\% and 7\%, respectively. Similar improvements are observed in terms of HD for all labeled data ratios. \modifR{Surprisingly, Co-training obtains a lower mean Dice than the partial supervision baseline for these two datasets. This is due to the fact that the final prediction of this method is the average of two separate networks. For challenging segmentation datasets like Prostate and Spleen, and when having few labeled images, it can happen that one of the two networks gives considerably worse predictions than the other. In this case, the poorly-performing network will hurt the co-training of both models, and the average of predictions will be worse than the prediction of a single network (e.g., Baseline). By exchanging only confident predictions in a self-paced manner, our method can effectively avoid this issue.}

To determine whether the improvements achieved by our method are significant, we ran a one-sided paired t-test for each segmentation task and performance metric. In Tables \ref{tab:ACDC_dice}-\ref{tab:spleen}, we underline the score of the best-performing method if it is significantly better (i.e., higher for DSC, smaller for HD) than the second-best one for the same setting. Significance is established when $\mr{p}\!<\!0.05$. It can be seen that our method yields significant improvements in all but 5 of the 20 test cases (i.e., 5 segmentation tasks $\times$ 2 labeled data ratios $\times$ 2 performance metrics).  

\begin{figure*}[!t]
\centering
\setlength{\tabcolsep}{0.5pt}
\renewcommand{\arraystretch}{1}
\begin{small}
\begin{tabular}{ccccccccc}
\viz{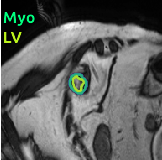} & \viz{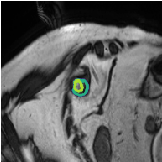} & \viz{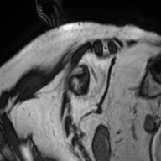} & \viz{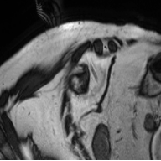} & \viz{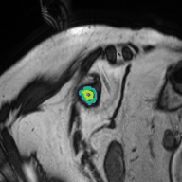} & \viz{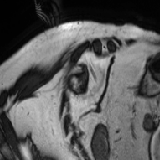} & \viz{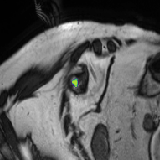} & \viz{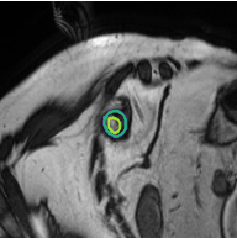} & \viz{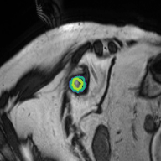} \\
\viz{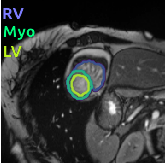} & \viz{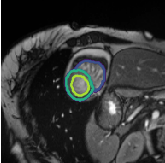} & \viz{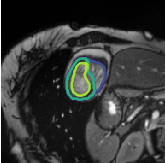} & \viz{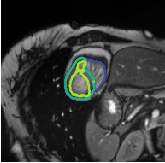} & \viz{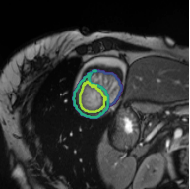} & \viz{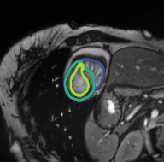} & \viz{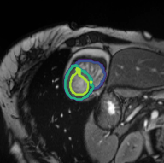} & \viz{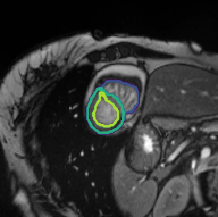} & \viz{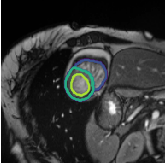} \\
\viz{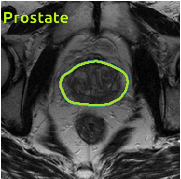} & \viz{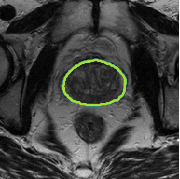} & \viz{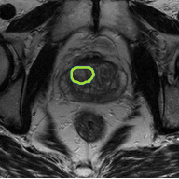} & \viz{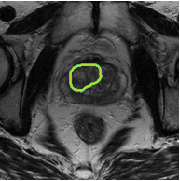} & \viz{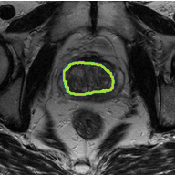} & \viz{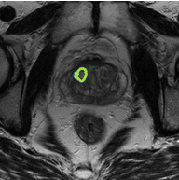} & \viz{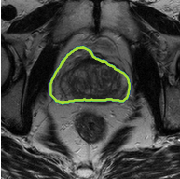} & \viz{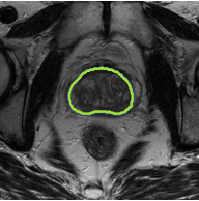} & \viz{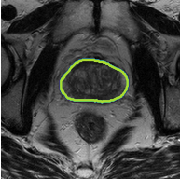} \\
\viz{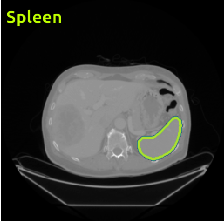} & \viz{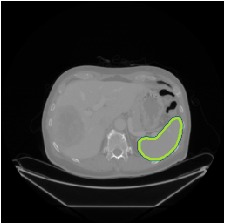} & \viz{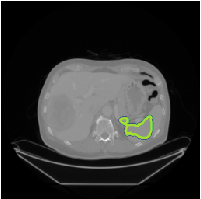} & \viz{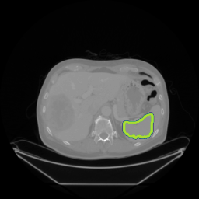} & \viz{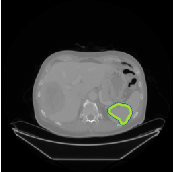} & \viz{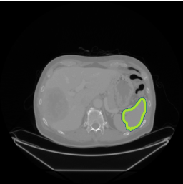} & \viz{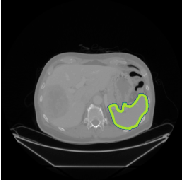} & \viz{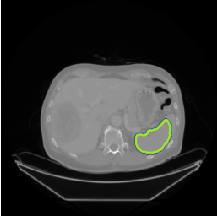} & \viz{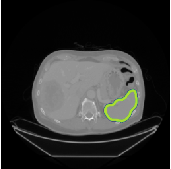} \\
\viz{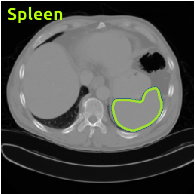} & \viz{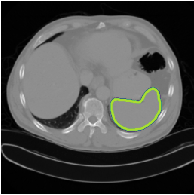} & \viz{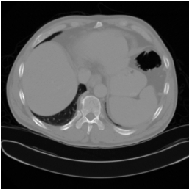} & \viz{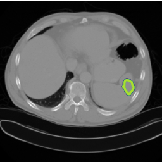} & \viz{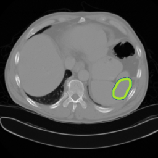} & \viz{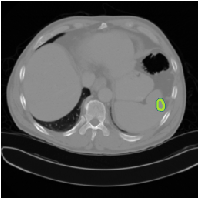} & \viz{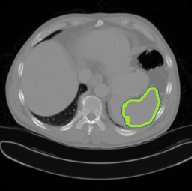} & \viz{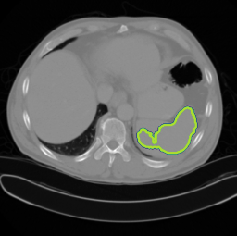} & \viz{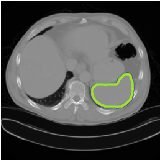} \\
Ground-truth & Full & Partial & Entropy min & DAN & Co-training & MT & UA-MT & Ours
\end{tabular}
\end{small}
\caption{Visual comparison of tested methods on test images. \textbf{Top two rows}: ACDC dataset. \textbf{Middle row}: Prostate dataset. \textbf{Bottom two rows}: Spleen dataset. A labeled data ratio of 10\% was used for all three datasets. Our method and Co-training were trained in a dual-view setting.}
\label{fig:visualization}
\end{figure*}

\begin{figure*}[!t]
\centering
\setlength{\tabcolsep}{0.5pt}
\renewcommand{\arraystretch}{1}
\begin{small}
\begin{tabular}{cccccc}
 \entropyMaps{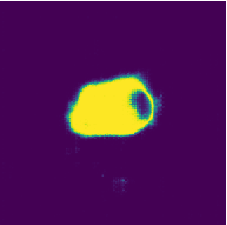} & \entropyMaps{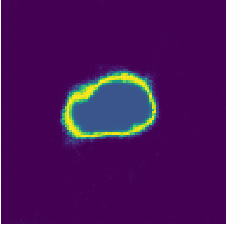} & \entropyMaps{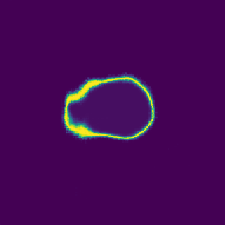} & \entropyMaps{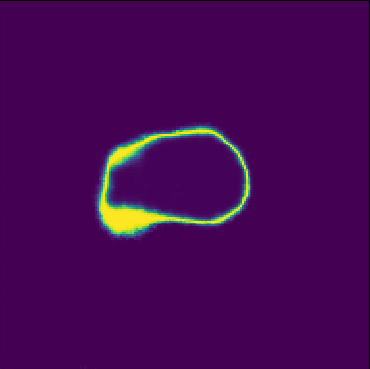} & \entropyMaps{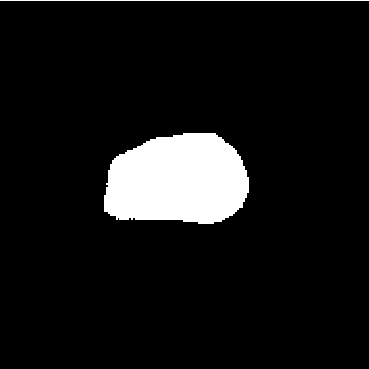} & \entropyMaps{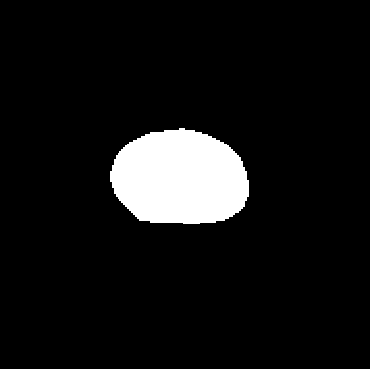}\\
\entropyMaps{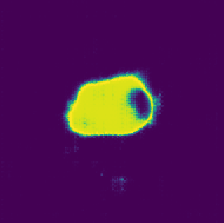} & \entropyMaps{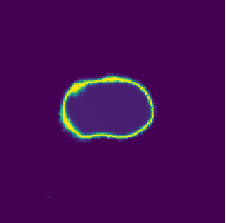} & \entropyMaps{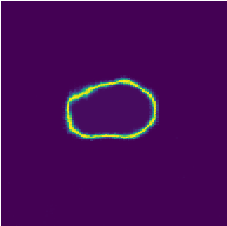} & \entropyMaps{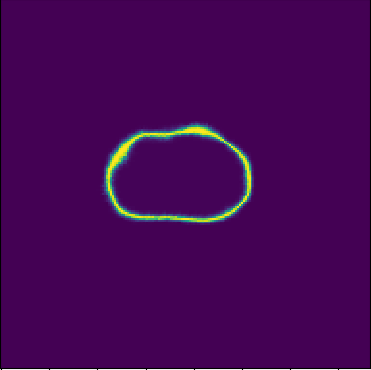} & \entropyMaps{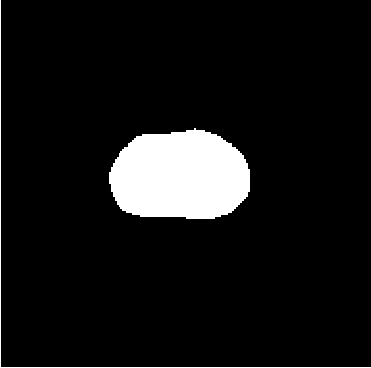} & \entropyMaps{p_target}\\
20 epochs & 40 epochs & 50 epochs & 80 epochs & Prediction & Ground-truth
\end{tabular}
\end{small}
\caption{Entropy maps, predicted segmentation and ground-truth mask for an image in the Prostate dataset. \textbf{Top row}: without our $\alpha$-entropy JSD loss. \textbf{Bottom row}: with the loss. It can be seen that the prediction becomes confident when using the proposed loss.}
\label{fig:entropy}
\end{figure*}

\subsection{Visualization of results}
We also confirm the effectiveness of our method by visually comparing segmentation results of tested approaches. Fig. \ref{fig:visualization} shows examples of results for test images in the three datasets, when training with 10\% labeled data. It can be seen that our method gives smoother segmentation contours compared to standard Co-training, MT, and UA-MT, and achieves segmentation predictions closer to the ground-truth mask, despite the low contrast of input images.

To visualize the impact of the uncertainty regularizer in the proposed method, Fig. \ref{fig:entropy} plots the prediction entropy maps of our method and standard Co-training for a test image in the Prostate dataset, at different training epochs. Compared to Co-training, the proposed uncertainty-regularized method gives a more confident prediction during training, leading to an improved segmentation.

\begin{table*}[!t]
\centering
\caption{Mean DSC (\%) of our method with different ablation settings on ACDC, Prostate and Spleen, using 5\%, 5\% and 7\% of labeled examples, respectively, for these datasets.}
\label{tab:ablation}
\setlength{\tabcolsep}{5pt}
\renewcommand{\arraystretch}{1}
\resizebox{\textwidth}{!}{
\begin{tabular}{cccccccc}
\toprule
  \B \multirow{2}{*}[-3pt]{\shortstack{Self-\\consistency}}& \B \multirow{2}{*}[-3pt]{\shortstack{Self-paced \\ learning}} & \multicolumn{4}{c}{\B ACDC} & \B \multirow{2}{*}[-3pt]{\shortstack{Prostate}} & \B \multirow{2}{*}[-3pt]{\shortstack{Spleen}} \\
 \cmidrule(l{6pt}r{6pt}){3-6}
 \B & \B & \B RV & \B Myo & \B LV & \B Mean\\
\midrule
 \xmark & \xmark & 76.28 \,(1.33) & 79.48 \,(0.61) & 90.95 \,(0.34) & 82.24 \,(0.76) & 51.02 \,(1.62) & 58.77 \,(3.86) \\ 
 \xmark & \cmark & 77.87 \,(0.56) & 79.66 \,(0.02) & 91.34 \,(0.38) & 82.96 \,(0.32) & 53.22 \,(0.83) & 61.33 \,(1.93) \\ 
 \cmark & \xmark & 82.05 \,(0.69) & 84.27 \,(0.59) & 92.20 \,(0.68) & 86.17 \,(0.65)& 68.09 \,(1.91) & 83.07 \,(2.89) \\ 
 \cmark & \cmark & \B 82.33 \,(0.16) & \B 84.46 \,(0.22) & \B \signif{92.47} \,(0.14) & \B 86.42 \,(0.17) & \signif{\B 70.15} \,(0.27) & \B \signif{85.50} \,(3.57) \\ 
\bottomrule
\end{tabular}
}
\end{table*}

\begin{table*}
\centering
\caption{Mean DSC (\%) of co-training methods on the ACDC dataset with $10\%$ labeled data, for 2 or 3 views.}
\label{tab:mutiview}
\setlength{\tabcolsep}{6pt}
\renewcommand{\arraystretch}{1}
\begin{footnotesize}
\begin{tabular}{clcccc}
\toprule
\multirow[b]{2}{*}{\B Views} & \multirow[b]{2}{*}{\B Method} & \multicolumn{4}{c}{\B ACDC}\\
\cmidrule(l{6pt}r{6pt}){3-6}
 & & \B RV & \B Myo & \B LV & \B Mean \\
\midrule
\multirow{2}{*}{2} 
& Co-training & 80.88 \,(0.53) & 83.69 \,(0.32) & 92.84 \,(0.31) & 85.80 \,(0.39) \\
& Ours & \B  \signif{83.85} \,(0.51) & \B \signif{86.42} \,(0.29) & \B \signif{93.06} (\,0.07) & \B \signif{87.78} \,(0.29) \\
\midrule
\multirow{2}{*}{3} 
& Co-training & 81.51 \,(0.19) & 84.49 \,(0.27) & 92.84 \,(0.40) & 86.28 \,(0.29) \\ 
& Ours &  \signif{\B 83.97} \,(0.58) &  \signif{\B 86.52} \,(0.42) &  \signif{\B 93.22} \,(0.11) &  \signif{\B 87.90 \,(0.37)} \\ 
\bottomrule
\end{tabular}
\end{footnotesize}
\end{table*}

\subsection{Ablation study}
To assess the respective contribution of the self-paced learning loss ($\Loss_{\mr{\textsc{spc}}}$) and self-consistency regularization loss ($\Loss_{\mr{reg}}$) in our co-training method, we performed an ablation study where we disable one or both of these losses during training. We carried out this study on the ACDC, Prostate, and Spleen datasets with labeled data ratios 5\%, 5\%, and 7\% respectively, and show the results in Table \ref{tab:ablation}.

We see that both the self-paced learning and self-consistency strategies improve performance when used by themselves. Hence, the self-paced learning strategy brings improvements of 0.7\%, 2.2\% and 2.6\% in mean DSC for ACDC, Prostate and Spleen, respectively, while self-consistency alone boosts DSC performance by 3.9\%, 17.1\% and 24.3\% for these datasets. However, combining both strategies results in even greater improvements of 4.2\%, 19.1\% and 26.7\%, demonstrating their synergy and complementary benefit.

\begin{table*}[!t]
\centering
\caption{Mean DSC (\%) of Mean Teacher and our methods on the ACDC dataset with $10\%$ labeled data and different backbone network architectures.}
\label{tab:diffbackbone}
\setlength{\tabcolsep}{6pt}
\renewcommand{\arraystretch}{1}
\begin{footnotesize}
\begin{tabular}{cp{66pt}cccc}
\toprule
\multirow[b]{2}{*}{\B Architectures} & \multirow[b]{2}{*}{\B Method} & \multicolumn{4}{c}{\B ACDC}\\
\cmidrule(l{6pt}r{6pt}){3-6}
 & & \B RV & \B Myo & \B LV & \B Mean \\
\midrule
\multirow{4}{*}{Enet} 
& Baseline & 77.51 \,(0.87) & 81.56 \,(0.46) & 91.72 \,(0.24) & 83.60 \,(0.53) \\
& MT & 82.91 \,(1.55) & 85.35 \,(0.57) & 92.37 \,(0.32) & 86.88 \,(0.81) \\
&UA-MT & 
83.63 \,(0.89) & 85.78 \,(0.16) & 92.13 \,(0.56) &  87.18 \,(0.54) \\ 
& Ours & \B 83.85 \,(0.51) & \B \signif{86.42} \,(0.29) & \B \signif{93.06} (\,0.07) & \B \signif{87.78} \,(0.29) \\
\midrule
\multirow{4}{*}{U-Net} 
& Baseline & 68.28 \,(1.61) & 79.94 \,(1.00) & 86.41 \,(0.29) & 78.21 \,(0.89) \\
& MT & 74.62 \,(1.10) & 80.66 \,(0.61) & 86.75 \,(0.27) & 80.68 \,(0.41) \\
&UA-MT & 
74.75 \,(0.88) & 80.92 \,(0.48) & 87.03 \,(0.43) &  80.90 \,(0.60) \\ 
& Ours & \B \signif{76.04} \,(0.73) & \B \signif{81.87} \,(0.22) & \B \signif{88.17} \,(0.18) & \B \signif{82.03} \,(0.38) \\
\midrule
\multirow{4}{*}{SegNet} 
& Baseline & 71.82 \,(0.82) & 79.84 \,(1.19) & 89.86 \,(1.49) & 80.51 \,(1.17) \\
& MT & 80.68 \,(0.46) & 85.17 \,(0.45) & 93.20 \,(0.14) & 86.35 \,(0.35) \\
&UA-MT & 
79.92 \,(0.22) & 85.24 \,(0.19) & 93.20 \,(0.27) &  86.12 \,(0.23) \\ 
& Ours & \B \signif{82.47} \,(0.45) & \B \signif{86.86} \,(0.27) & \B 93.50 \,(0.33) & \B \signif{87.61} \,(0.35) \\
\bottomrule
\end{tabular}
\end{footnotesize}
\end{table*}

\begin{table}[!t]
\centering
\caption{Training and inference time of the tested methods, for a batch size of 1.}
\label{tab:computation}
\setlength{\tabcolsep}{6pt}
\renewcommand{\arraystretch}{1}
\begin{footnotesize}
\begin{tabular}{lccc}
\toprule
\B Method & \B Views & \B \shortstack{Training time \\ (ms\,/\,batch)} & \B \shortstack{Inference time \\ (ms\,/\,batch)} \\
\midrule
Baseline & 1 & 285 & 87\\
Entropy min & 1 & 310 & 87 \\
DAN & 1 & 450 & 87 \\
MT& 1 & 380 & 87 \\
\modifR {UA-MT} & \modifR{1} & \modifR{395} & \modifR{87}\\
\midrule
\multirow{2}{*}{Co-training} & 2 & 540 & 128 \\
& 3 & 760 & 248 \\
\midrule 
\multirow{2}{*}{Ours} & 2 & 660 & 128\\
& 3 & 990 & 248\\
\bottomrule
\end{tabular}
\end{footnotesize}
\end{table}

\subsection{Multi-view analysis}
In previous experiments, we tested our co-training method in a dual-view setting where two segmentation networks are trained in a collaborative manner. In the next analysis, we assess whether increasing the number of views can further improve segmentation performance. Table \ref{tab:mutiview} compares the results of our method trained with 10\% of labeled examples from the ACDC dataset, using either 2 or 3 views. We see that jointly training 3 segmentation networks instead of 2 boosts performance for both standard Co-training and our proposed method, bringing the mean DSC even closer to that of full-supervision. However, further experiments showed that using more than 3 models leads to negligible improvements while requiring more computational resources. 

\subsection{Impact of network architecture}
To evaluate the robustness of the proposed method to different backbone architectures, we also employed U-Net~\citep{unet2015} and SegNet~\citep{segnet2015} as the underlying segmentation network. The U-Net architecture consists of a contracting path and an expansive path. The contracting path follows the architecture of a convolutional network, which consists of the repeated application of two 3$\times$3 convolutions, each followed by a rectified linear unit (ReLU) and a max pooling operation for downsampling. Every step in the expansive path consists of an upsampling of the feature map followed by a convolution that halves the number of feature channels, a concatenation with the correspondingly cropped feature map from the contracting path, and two 3$\times$3 convolutions followed by a ReLU. At the final layer, a 1$\times$1 convolution is used to map each feature vector to the desired number of classes. The number of U-Net architecture parameters is 31.04\,M. Finally, the SegNet architecture has an encoder and a corresponding decoder, followed by a final pixel-wise classification layer. The encoder consists of 13 convolutional layers identical to the first 13 layers in the VGG16 network. Each encoder layer has a corresponding decoder layer. The final decoder output is fed to a multi-class soft-max classifier to produce class probabilities for each pixel. SegNet has a total of 29.46\,M trainable parameters.

\modifR{The DSC performance of our method, Mean Teacher and UA-MT obtained with different backbone architectures, for the ACDC data with 10\% of labeled data, is given in Table \ref{tab:diffbackbone}. Results show our method to provide consistently-higher accuracy than UA-MT for all backbone networks. Comparing the segmentation architectures to each other, we find that ENet and SegNet yield a similar mean DSC, which is about 5.6\% greater than U-Net. Considering that ENet requires much less computation and memory than SegNet, we conclude that this architecture is best for our method.}

Moreover, to evaluate the runtime efficiency of our method, we provide in Table \ref{tab:computation} the average training time and inference time of tested approaches using ENet as backbone network and a batch size of 1. The baseline model needs to compute only a single loss per pass, thus has the lowest training times. Although Mean Teacher uses two networks in training, only the Student model is updated via back-propagation (the Teacher's parameters are updated with EMA of the Student's). As Mean Teacher, dual-view Co-training also requires training two models. However, the parameters of these models can be updated in parallel, instead of sequentially like Mean Teacher. Because it combines self-ensembling and self-paced co-training, our method requires more computations than Co-training, resulting in a 22\% longer training time than this approach in the dual-view setting. In terms of the inference time, both our method and Co-training need to do a forward pass on two separate networks, which increases computations compared to other approaches. As in training, this could also be done in parallel to speed-up inference.

\section{Conclusion}\label{sec:conclusion}

We proposed a self-paced and self-consistent co-training method for semi-supervised image segmentation. Our method extends standard co-training by focusing first on easier regions of unlabeled images, and by encouraging both consistency and confidence across the different models during training. Our self-paced learning strategy uses a end-to-end differentiable loss based on generalized JSD to dynamically control the importance of individual pixels on co-training the different segmentation networks. Moreover, a self-consistency loss based on temporal ensembling is used to further regularize the training of individual models and improve performance when annotated data is very limited. We evaluated the potential of our method in three challenging segmentation tasks, including images of different modalities. Experimental results showed our proposed method to outperform state-of-art approaches for semi-supervised segmentation and yield a performance close to full-supervision while using only a small fraction of the labeled data.

A limitation of the proposed method is the need to run multiple segmentation networks, which increases the computational requirements. Although parallel computation techniques can be adopted to speed up training and inference, this limitation could be addressed alternatively by creating a single model that distillates the knowledge of individual models across views, similar to \citep{tarvainen2017mean}. One could also reduce training and inference times by having the co-trained networks share some of their layers, for example allowing only the last few layers of the decoder to differ.
Another potential drawback of our method is the need to balance different loss terms that may compete against one another during training. To alleviate this problem, a useful extension of this work could be to investigate self-tuning mechanisms which can adapt more efficiently to new datasets. As future work, we plan to extend our method to the segmentation of 3D and multi-modal images. We will also investigate other strategies for self-paced learning and self-consistency in our co-training framework. 

\section*{Acknowledgments}
This work was supported in part by the Natural Sciences and Engineering Research Council of Canada (NSERC) Discovery Grants Program under grant RGPIN-2018-05715.

\bibliographystyle{model2-names.bst}
%\biboptions{authoryear}
\bibliography{refs}

\appendix

\section{Proof of Theorem 1}\label{sec:proof1}
\begin{proof}
With fixed $\{\params^k\}$ and $\{\vec{\pslab}_{ui}\}$, the optimal learning weights $w_{uik}$ corresponding to pixel $i$ of unlabeled example $u$ and model $f^k$ are found by solving
\begin{equation}
\min_{w_{uik} \in [0,1]} \ \frac{\gamma}{2}w^2_{uik} \, + \, \big(\kl{\pp^k_{ui}}{\vec{\pslab}_{ui}} - \gamma\big)\,w_{uik}
\end{equation}
If $\kl{\pp^k_{ui}}{\vec{\pslab}_{ui}} \geq \gamma$, since $w_{uik} \geq 0$, the minimum is obviously achieved for $w^*_{uik}=0$. Else, if $\kl{\pp^k_{ui}}{\vec{\pslab}_{ui}} < \gamma$, we find the optimal weight by deriving the function w.r.t. $w_{uik}$ and setting the result to zero, which gives
\begin{equation}
w^*_{uik} \ = \ 1 \, - \, \frac{1}{\gamma}\kl{\pp^k_{ui}}{\vec{\pslab}_{ui}}.
\end{equation}
Since both $\gamma$ and $\kl{\pp^k_{ui}}{\vec{\pslab}_{ui}}$ are non-negative, we have that $w^*_{uik} \in [0,1]$, thus it is a valid solution. Considering both cases simultaneously, we therefore get 
\begin{equation}
w^*_{uik} \ = \ \max\Big(1 - \frac{1}{\gamma}\kl{\pp^k_{ui}}{\vec{\pslab}_{ui}}, \, 0\Big)
\end{equation}
\end{proof}

\section{Proof of Theorem 2}\label{sec:proof2}
\begin{proof}
Considering learning weights $\{\ww_{ui}\}$ as fixed, optimizing model parameters $\{\params^k\}$ and pseudo-labels $\{\pslab_{ui}\}$ in (\ref{eq:spct}) corresponds to:
\begin{align}\label{update-pseudo}
 & \min_{\{\params^k\}, \{\vec{\pslab}_{ui}\}} \ \frac{1}{|\unlab|}\sum_{\img_u \in \unlab} \sum_{k=1}^K \sum_{i \in \pixels}
 w_{uik} \, \kl{\pp^k_{ui}}{\vec{\pslab}_{ui}} \\ 
 & \quad \tx{s.t.} \ \sum_{j \in \classes} \pslab_{uij} \, = \, 1, \ \forall u, \forall i; \ \ \pslab_{uij} \in [0,1], \, \forall u,\forall i, \forall j.\nonumber
\end{align}
Since the pseudo-labels $\vec{\pslab}_{ui}$ for each pixel are decoupled in the loss, we can optimize them independently. For each resulting sub-problem, we deal with the constraint that $\vec{\pslab}_{ui}$ is a probability distribution with a Lagrangian formulation and convert the problem into
\begin{align}
 & \max_{\mu} \, \min_{\vec{\pslab}_{ui}} \ \sum_{k=1}^K
 w_{uik} \, \kl{\pp^k_{ui}}{\vec{\pslab}_{ui}} \ - \ \mu\Big(\sum_{j \in \classes}\pslab_{uij} \, - \, 1\Big)\nonumber\\
 & \ = \ -\sum_{k=1}^K \sum_{j \in \classes} w_{uik} \, p^k_{uij} \log \frac{\pslab_{uij}}{p^k_{uij}} 
 \, - \, \mu\Big(\sum_{j \in \classes}\pslab_{uij} \, - \, 1\Big)
\end{align}
where $\mu$ is the Lagrange multiplier corresponding to the one-sum constraint on $\vec{\pslab}_{ui}$. Next, we derive this function with respect to each $\pslab_{uij}$ and set the result to zero, yielding
\begin{align}\label{eq:pseudol-def}
 \pslab^*_{uij} \ = \ -\frac{1}{\mu}\sum_{k=1}^K w_{uik} \, p^k_{uij}
\end{align}
To find $\mu$ we use the constraint that $\sum_j \pslab_{uij}=1$:
\begin{align}
& - \frac{1}{\mu}\sum_{j \in \classes} \sum_{k=1}^K w_{uik} \, p^k_{uij} \ = \ 1\nonumber\\
& \mu \ = \ -\sum_{j \in \classes} \sum_{k=1}^K w_{uik} \, p^k_{uij} \ = \ -\sum_{k=1}^K w_{uik}
\end{align}
Using this equation in (\ref{eq:pseudol-def}) thus yields
\begin{align}\label{eq:pslab}
\pslab^*_{uij} & \ = \ \frac{\sum_{k=1}^K w_{uik} \, p^k_{uij}}{\sum_{k=1}^K w_{uik}}
\ = \ \sum_{k=1}^K \pi_{uik} \, p^k_{uij}.
\end{align}
We finally insert (\ref{eq:pslab}) in the loss of (\ref{update-pseudo}) to get
\begin{align}
& \min_{\{\params^k\}, \{\vec{\pslab}_{ui}\}} \ -\frac{1}{|\unlab|}\sum_{\img_u \in \unlab} \sum_{k=1}^K \sum_{j \in \classes}
 w_{uik} \, p^k_{uij} \, \log \frac{\pslab_{uij}}{p^k_{uij}} \nonumber\\
 & \ = \ \frac{1}{|\unlab|}\sum_{\img_u \in \unlab} \sum_{i \in \pixels} \rho_{ui} \Big[\!\!-\!\sum_{j \in \classes}\!\big(\sum_{k=1}^K\!\pi_{uik} \, p^k_{uij}\big) \log \big(\sum_{k'=1}^K\!\pi_{uik'} \, p^k_{uij} \big) \nonumber\\
  & \quad \ + \ \sum_{k=1}^K \pi_{uik} \sum_{j \in \classes} p^k_{uij} \log \, p^k_{uij}\Big] \nonumber\\
  & \ = \ \frac{1}{|\unlab|}\sum_{\img_u \in \unlab} \sum_{i \in \pixels} \rho_{ui} \Big[ \entr\big(\sum_{k=1}^K \pi_{uik} \, \pp^k_{ui}\big) \, - \, \sum_{k=1}^K \pi_{uik} \entr\big(\pp^k_{ui}\big)\Big] \nonumber\\
  & \ = \ \frac{1}{|\unlab|}\sum_{\img_u \in \unlab} \sum_{i \in \pixels} \rho_{ui} \, \JSD_{\ppi_{ui}}\big(\pp^1_{ui}, \, \ldots, \pp^K_{ui}\big). 
\end{align}
\end{proof} 

\end{document}